\documentclass[10pt,journal,compsoc]{IEEEtran}

\ifCLASSOPTIONcompsoc
  \usepackage[nocompress]{cite}
\else
  \usepackage{cite}
\fi

\ifCLASSINFOpdf
\else
\fi

\usepackage{amsmath,amssymb} % define this before the line numbering.
\usepackage{color}
 \newcommand\figcaption{\def\@captype{figure}\caption}
 \newcommand\tabcaption{\def\@captype{table}\caption}
 \usepackage{subcaption}
\usepackage{graphicx}
\usepackage{amsmath}
\usepackage{amssymb}
\usepackage{booktabs}
\usepackage{textcomp}  % textdegree
\usepackage{amsfonts}
\usepackage{multirow}
\usepackage{algorithm}
\usepackage{algorithmic}
\usepackage{float}
\usepackage{tabularx}
\usepackage{array}
\usepackage[misc]{ifsym}
\usepackage[pagebackref=true,breaklinks=true,letterpaper=true,colorlinks,bookmarks=false]{hyperref}

\usepackage{subcaption}
\usepackage{lipsum}

\usepackage{xspace}

\usepackage[pagebackref=true,breaklinks=true,colorlinks,bookmarks=false]{hyperref}

\newcommand*{\eg}{\emph{e.g.}}
\newcommand*{\ie}{\emph{i.e.}}

\usepackage{ragged2e}

\begin{document}
\title{InvPT++: Inverted Pyramid Multi-Task Transformer for Visual Scene Understanding}

\author{Hanrong Ye and~Dan Xu % <-this % stops a space
\IEEEcompsocitemizethanks{
    \IEEEcompsocthanksitem Hanrong Ye and Dan Xu are with the Department of Computer Science and Engineering, The Hong Kong University of Science and Technology (HKUST), Hong Kong.\protect~E-mail: \{hyeae, danxu\}@cse.ust.hk}% <-this % stops an unwanted space
%\thanks{Manuscript received.}
}


% \markboth{Journal of \LaTeX\ Class Files,~Vol.~14, No.~8, August~2015}
% {Shell \MakeLowercase{\textit{et al.}}: Bare Demo of IEEEtran.cls for Computer Society Journals}

\IEEEtitleabstractindextext{%
\justifying
\begin{abstract}
Multi-task scene understanding aims to design models that can simultaneously predict several scene understanding tasks with one versatile model. 
Previous studies typically process multi-task features in a more local way, and thus cannot effectively learn spatially global  and cross-task interactions, which hampers the models' ability to fully leverage the consistency of various tasks in multi-task learning.
To tackle this problem, we propose an Inverted Pyramid multi-task Transformer, capable of modeling cross-task interaction among spatial features of different tasks in a global context.
Specifically, we first utilize a transformer encoder to capture task-generic features for all tasks. 
And then, we design a transformer decoder to establish spatial and cross-task interaction globally, and a novel UP-Transformer block is devised to increase the resolutions of multi-task features gradually and establish cross-task interaction at different scales. Furthermore, two types of Cross-Scale Self-Attention modules, \emph{i.e.}, Fusion Attention and Selective Attention, are proposed to efficiently facilitate cross-task interaction across different feature scales.
An Encoder Feature Aggregation strategy is further introduced to better model multi-scale information in the decoder.
Comprehensive experiments on several 2D/3D multi-task benchmarks clearly demonstrate our proposal's effectiveness, establishing significant state-of-the-art performances. 
The codes are publicly available  \href{https://github.com/prismformore/Multi-Task-Transformer/tree/main/InvPT}{here}.
\end{abstract}

% Note that keywords are not normally used for peer review papers.
}

\maketitle

\IEEEdisplaynontitleabstractindextext

% \fi
\IEEEpeerreviewmaketitle

\section{Introduction}
The concurrent learning and reasoning of multiple interrelated tasks is the goal of multi-task scene understanding~\cite{mtlsurvey}. This technique is applicable to various AI products including robots, self-driving vehicles, and augmented reality~(AR) headsets.
A majority of visual scene understanding tasks (\emph{e.g.}, segmentation, depth estimation, and object detection) require a pixel-level understanding of visual scenes~\cite{astmt}. These tasks, which involve dense predictions, inherently possess considerable explicit and implicit correlations at the pixel level~\cite{padnet}. 
These correlations, when fully exploited, can significantly boost the overall performance of multi-task models. Furthermore, the simultaneous learning and prediction of multiple tasks inherently offer greater efficiency than the separate training and inference of several single-task models, as various tasks can share numerous network modules.
Such observations have sparked a growing interest among researchers to develop advanced multi-task learning methods from different perspectives, aiming to more effectively tackle the complexities of multi-task scene understanding.

\begin{figure}[!t]
    \centering
    \includegraphics[width=\linewidth]{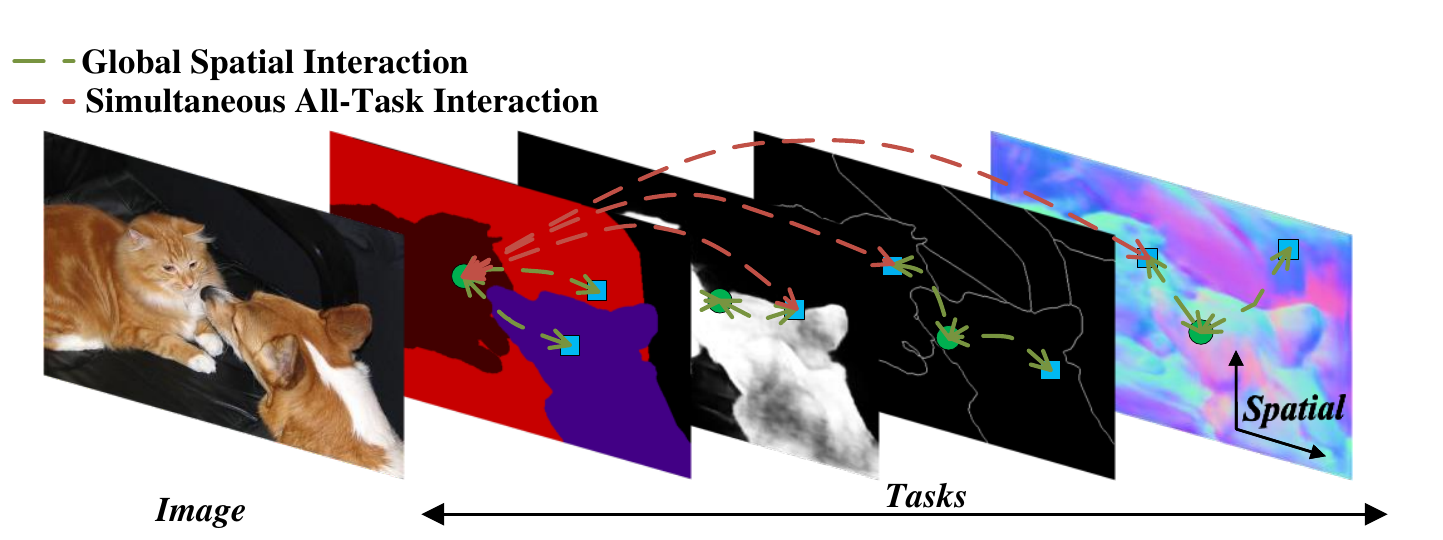}
    \vspace{-8pt}
    \caption{Effective information exchange among tasks is particularly important for multi-task dense scene understanding, InvPT++ proposes a pioneering approach to construct fully-connected global modeling of different task features along spatial and task dimensions in an efficient manner.}
    \label{fig:multi-task contexts}
\end{figure}

The prevailing approaches for multi-task scene understanding primarily depend on Convolutional Neural Networks (CNNs), with considerable progress made by devising multi-task optimization loss functions~\cite{kendall2018multi}, as well as creating~\cite{padnet} or exploring~\cite{gao2020mtl} multi-task network architectures. While these methods have demonstrated encouraging results, they remain constrained by the intrinsic properties of convolution kernels extensively employed in their deep learning frameworks, \ie, locality as discussed in prior works~\cite{NonLocal2018,attn_aug}. These kernels model essential spatial and task-related contexts within relatively localized receptive fields.
Attention-based techniques have been proposed to address this limitation~\cite{papnet,psd,atrc}, but the extent of cross-task interaction they enable is still narrow. Thus, capturing global dependencies and relationships of all tasks, which is vital for multi-task learning in dense scene understanding, remains an open issue.

Recently, transformer models have been introduced to capture long-range spatial relationships in dense prediction problems~\cite{dpt,hrformer}. However, these models primarily focus on single-task learning settings, without an exploration of joint modeling for multiple dense prediction tasks using transformers. Developing a unified transformer framework that globally models both spatial and cross-task correlations is not straightforward and challenging. Existing transformers face difficulty when adopting higher feature resolution because of the exponentially larger  computational complexity, therefore many methods choose to  drastically reduce the spatial resolution to mitigate computational overhead~\cite{vit,pvt}.
However, this limits the performance of dense prediction tasks whose performances are significantly influenced by the resolution of the feature maps generated by the model. 
Simultaneous to our work, several studies have investigated multi-task transformers~\cite{bhattacharjee2022mult,xu2022mtformer,xu2022multi}. However, their attention maps still possess limited context and lack the capability to model spatial and cross-task interactions globally among features of different tasks.

To address the aforementioned challenges, we have developed an innovative end-to-end Inverted Pyramid Multi-task Transformer framework, abbreviated as ``InvPT++''. This framework effectively models long-range dependencies within spatial and all-task contexts while concurrently learning in high resolution, producing accurate dense prediction maps for multi-task dense scene understanding. InvPT++ comprises three primary components. As the first part of the model, an InvPT++ transformer encoder derives generic visual representations from input images via spatial long-range relationship modeling. 
Following this, an InvPT++ decoder, equipped with several innovative UP-Transformer blocks, refines multi-task features at progressively increasing resolutions. More importantly, it is capable of capturing cross-task interaction among various tasks at different scales within a global and spatial context.
We further propose a Cross-Scale Self-Attention, which is able to utilize the attention message from the previous transformer layer to help model cross-task interaction at the current layer, in the designed UP-Transformer blocks.
We develop two types of Cross-Scale Self-Attention: Fusion Attention and Selective Attention.
The Fusion Attention learns a pair of weights to fuse the attention message from the previous layer and the attention score at the current layer.
Differently, Selective Attention is designed to further boost computational efficiency. It selects a subset of important tokens from the original token sequence and employs only these important tokens in self-attention calculations. The process of identifying the important tokens is guided by the attention scores of the previous layer. 
Additionally, we devise an Encoder Feature Aggregation (EFA) strategy to augment decoder features by incorporating multi-scale features obtained from the InvPT++ encoder.
InvPT++ clearly outperforms previous state-of-the-art methods on several commonly used benchmarks for multi-task dense scene understanding, \emph{i.e.}, PASCAL-Context, NYUD-v2, Cityscapes, and Cityscapes-3D datasets.
The proposed modules are carefully examined by the experimental results on these competitive benchmarks.

\par Compared with our previous conference version InvPT~\cite{invpt2022}, in this work, we further introduce a powerful Selective Attention strategy into the Cross-Scale Self-Attention module, which greatly reduces redundant computation in the self-attention module by leveraging cross-scale attention information from the previous layer. 
The Selective Attention mechanism enables InvPT++ to discover important tokens before the computation of self-attention maps and use only the important tokens in self-attention computation. The selection of important tokens is effectively guided by the attention message from the previous layer.
This innovation enables InvPT++ to markedly surpass InvPT while utilizing 22.51\% fewer FLOPs in the transformer decoder. We remarkably extend our experiments to investigate the effectiveness of InvPT++ by considering two new and more challenging 2D/3D multi-task scene understanding benchmarks, \emph{i.e.}, the Cityscapes~\cite{Cordts2016Cityscapes} and Cityscapes-3D~\cite{gahlert2020cityscapes3d} datasets. Substantial results and analysis in qualitative and quantitative aspects are further provided in the experiments.

The contributions of this work are three-fold:
\begin{itemize}
\setlength{\parskip}{-1pt}
  \item 
  A powerful transformer model ``InvPT++'', designed for multi-task visual scene understanding, is proposed. To the best of our knowledge, it is the first multi-task deep framework that enables a spatially global context in the modeling of cross-task interaction, which empowers the tasks to mutually enhance their learning within a single model.
  \item 
  An UP-Transformer block is developed, which effectively facilitates the cross-task interaction and multi-scale learning of multi-task features. By stacking multiple UP-Transformer blocks, we construct the InvPT++ decoder that can generate high-resolution task-specific features for different tasks. 
  \item
  Inside the UP-Transformer block, we propose two types of Cross-Scale Self-Attention modules: Fusion Attention and Selective Attention. They both help model cross-task interaction by passing attention information across different scales. An Encoder Feature Aggregation method is further designed to help the learning of multi-scale features in the multi-task decoder.
\end{itemize}

The effectiveness of different components of the proposed InvPT++ is extensively validated, and InvPT++ also produces new state-of-the-art performances on several major 2D/3D multi-task dense scene understanding benchmarks, including PASCAL-Context, NYUD-v2, Cityscapes, and Cityscapes-3D datasets. 
The rest of the paper is organized as follows: Section~\ref{sec:relatedwork}
reviews the literature in related fields. 
Section~\ref{sec:method} gives a detailed introduction to different modules in the proposed InvPT++.
Section~\ref{sec:experiments} demonstrates our experimental results to verify the effectiveness of our proposal.
Section~\ref{sec:conclusion} concludes the paper.

\section{Related Work}
\label{sec:relatedwork}
The most related works are organized into two aspects, \emph{i.e.}, multi-task learning for dense scene understanding, and visual transformers.

\par\noindent\textbf{Multi-task Learning for Dense Scene Understanding} 
Multi-task learning has two-fold strengths compared with single-task learning: (i) Firstly, jointly learning multiple tasks in one model is naturally more efficient than learning tasks with different models separately, since different tasks can share some common information captured by task-shared network modules.
(ii) Secondly, as different tasks can help each other via information complementation, they can outperform single-task models with a proper design of multi-task models. 
Given the significant commonalities shared among different dense scene understanding tasks, the strengths of multi-task learning have inspired a substantial amount of research in this area.

\begin{figure*}[!t]
	\centering
	\includegraphics[width=1\textwidth]{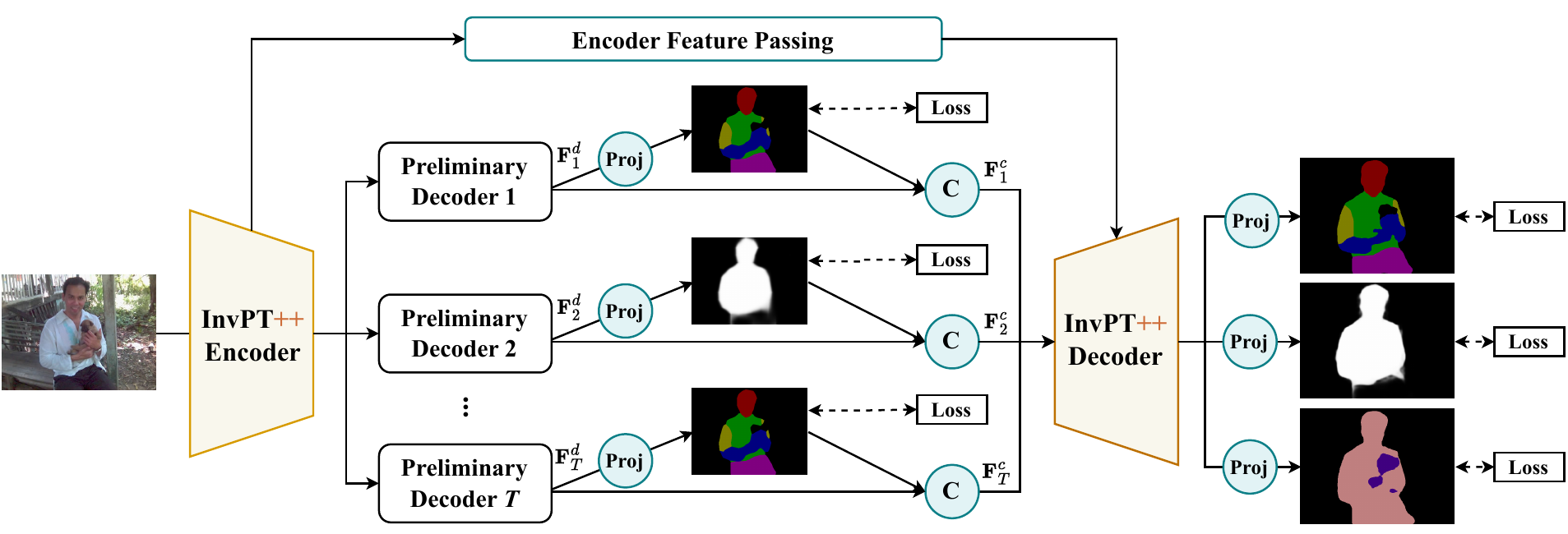}
\caption{
Illustration of our InvPT++ framework.
The process comprises three parts:
(i) The InvPT++ transformer encoder, shared among all tasks, acquires task-agnostic visual features from the input image.
(ii) A set of task-specific preliminary decoders generate task-specif feature $\mathbf{F}^d_t$ and preliminary predictions $\mathbf{P}_t$ for each task $t$, which are subsequently concatenated on the channel dimension to form the combined feature $\mathbf{F}^c_t$.
(iii) A unique InvPT++ transformer decoder processes $\mathbf{F}^c_t$ of all $T$ tasks as input, enables global cross-task interaction in spatial and task dimensions, and refines task features with incrementally increasing resolutions.
In the figure, \textcircled{c} represents concatenation, and "Proj" refers to the linear projection layer.
}
\label{fig:framework}
\end{figure*}

Existing deep multi-task learning methods explore two different directions, \emph{i.e.}, optimization-based and architecture-based. Optimization-based methods concentrate on enhancing the multi-task optimization process, typically by designing multi-task balancing techniques for loss functions and gradients of different tasks~\cite{gradientsign,gradnorm,kendall2018multi}, or by enforcing cross-task consistency through contrastive learning regularization loss~\cite{contrastiveMTL2023,li2022learning}.
Conversely, architecture-based methods primarily focus on developing effective multi-task model architectures, either throughout the entire deep model~\cite{nddr,liu2019MTAN,crossstitch,sluice,nddr} or specifically in the decoder stage~\cite{padnet,papnet,psd,zhang2021transfer,mti,invpt2022}.

Significant development has been witnessed in the domain of multi-task dense scene understanding, as evidenced by the pioneering studies~\cite{kendall2018multi,astmt,crossstitch,nddr,padnet,kim2023vtm,taskprompter2023}. 
To design better multi-task model architecture, PAD-Net~\cite{padnet} proposed an effective cross-task interaction module named ``Multi-Modal Distillation'' enabling cross-task information fusion. 
MTI-Net~\cite{mti} improved PAD-Net by incorporating a multi-scale modeling strategy and further boosted the performance. 
However, these works can only model cross-task relationships in a highly spatially local manner because of the usage of CNNs in the cross-task interaction modules. 
Therefore, some researchers turn to self-attention~\cite{transformer} for modeling cross-task interaction.
\cite{zhang2021transfer}, \cite{papnet}, and \cite{psd} developed spatially global or local attention-based methods to learn task-specific representation for each task, and then propagate the features across the tasks. However, these works still model cross-task interaction in a local context which hampers the further improvement of the multi-task models.

Several concurrent works have incorporated self-attention in multi-task models. ATRC~\cite{atrc} utilizes the neural architecture search technique to identify suitable attention types for individual task pairs. However, it falls short in capturing a global context that encompasses all tasks. 
MTFormer~\cite{xu2022mtformer} and MQTransformer~\cite{xu2022multi} develop different cross-task information fusion modules, yet the multi-task information cannot interact with each other in a spatially global context. Similar to MTAN~\cite{liu2019MTAN}, the task-specific modules of MulT~\cite{bhattacharjee2022mult} directly query the task-shared backbone feature to acquire task-specific features but are unable to model cross-task interactions.
MultiMAE~\cite{bachmann2022multimae} learns a versatile backbone model but requires fine-tuning on each single task, making it essentially a series of single-task models instead of one multi-task model.
Unlike these researches on multi-task learning, we introduce a novel transformer framework in this paper that facilitates spatially global interaction across features of different tasks that crucially affects multi-task model performance.

\par\noindent\textbf{Visual Transformers}
The study of visual transformers is a blooming research area. Initially designed for NLP tasks~\cite{transformer}, transformers have been widely used in the study of computer problems~\cite{vit,bhojanapalli2021understanding,bai2021transformers}, including multi-modal learning~\cite{clip,dalle}, 2D~\cite{parmar2018image,vit,detr,ipt,han2021transformer,chen2021pix2seq} and 3D tasks~\cite{pan20213d,mao2021voxel,Zhao_2021_ICCV}.
Regarding the architecture design of visual transformers, previous works can be categorized into several classes:
(i) Exploit useful inductive bias with self-attention. For instance, Swin Transformer~\cite{swin} proposes to compute local attention instead of global attention based on shifted windows, to improve computational efficiency. Focal Transformer~\cite{focal} applies a multi-granularity strategy when defining the context in self-attention motivated by the belief that adjacent areas are more important. This helps reduce the size of the attention map and improve computational efficiency. 
(ii) Marrying transformers with CNNs~\cite{srinivas2021bottleneck,pvt,cvt}.
BOTNet~\cite{srinivas2021bottleneck} embeds a special multi-head self-attention module in some of the layers of ResNet.  PVT~\cite{pvt} and CVT~\cite{cvt}, conversely, improve transformer architecture with CNNs to increase performance and efficiency.
(iii) Inventing special training techniques for transformers. For example, DEIT~\cite{deit} devised a special learnable token for knowledge distillation, which makes it possible to learn from pre-trained CNNs. DRLOC~\cite{liu2021efficient} utilized an auxiliary task to help the transformer learn the relative location of tokens.
DPT~\cite{dpt} put forward a transformer designed for dense prediction tasks, utilizing transposed convolution layers to obtain high-resolution feature maps. HRFormer~\cite{hrformer} maintains multi-scale features across different layers of the models via using local attention to reduce overhead.

Based on our survey, this study is the first to concurrently model all-task interaction within a spatially global context at multiple scales using a unified transformer model for multi-task scene understanding, which helps it greatly outperform previous methods on several benchmarks. 

\section{Method}
\label{sec:method}
\subsection{Framework Overview}
The overall framework overview of the proposed InvPT++ is depicted in Fig.~\ref{fig:framework}. 
It can be divided into three parts: a task-generic InvPT++ transformer encoder, several task-specific preliminary decoders, and an InvPT++ transformer decoder. 
The transformer encoder is made to be shared for all the tasks, acquiring task-shared features from the same input image. Following this, the preliminary decoders extract task-specific features for all tasks, which are used to compute the preliminary predictions. The preliminary predictions are learned via the task labels and the loss functions. For each task, the corresponding task-specific features and preliminary predictions are concatenated on the channel dimension, forming the input for the subsequent InvPT++ decoder. The InvPT++ decoder is designed to model all-task interaction in a spatially global manner and refine the task-specific features of all tasks. Task-specific linear projection layers then calculate the final predictions from the refined task-specific features. The following sections will provide a detailed description of these modules.

\subsection{InvPT++ Encoder}
The InvPT++ encoder is a transformer encoder designed to extract generic task-shared features given the image. In each transformer layer, the self-attention module aids in learning a global feature representation by modeling long-range spatial dependencies among different spatial locations. The encoder produces a feature sequence that is then reshaped into a spatial feature map. The shape of the feature map is $H_0\times W_0$, where $H_0$ and $W_0$ represent the height and width of the feature map, respectively. 
Assuming we learn $T$ tasks, the task-generic feature obtained from the encoder is then passed to $T$ preliminary decoders to generate the corresponding $T$ feature maps. In the experiments, we explore different encoder choices, \emph{e.g.}, ViT~\cite{vit} with global attention, and Swin Transformer~\cite{swin} with window attention.

\begin{figure}[t]
\centering
\includegraphics[width=1\linewidth]{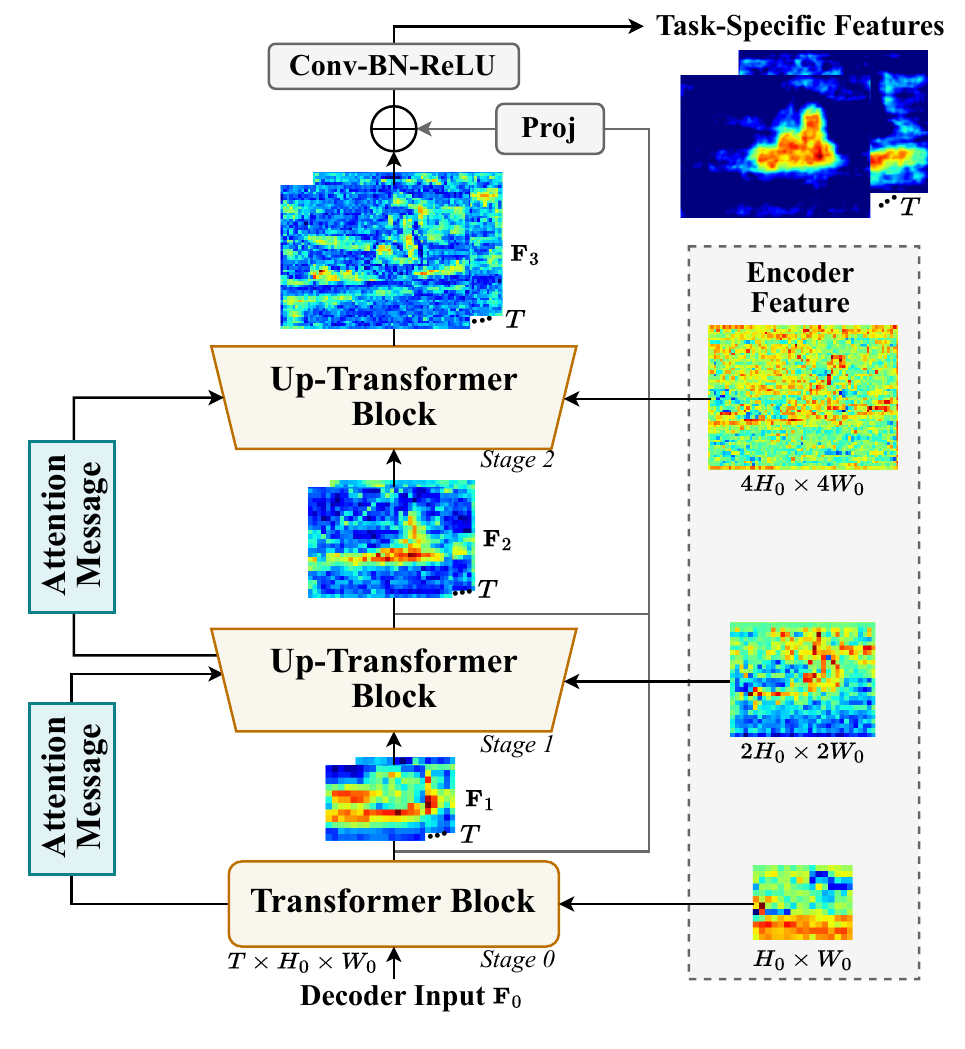}
\caption{
Design of the InvPT++ decoder. It is composed of three stages, each of which globally models the cross-task interaction between features of all tasks in a spatial global context at different scales via a specially designed transformer block.
The spatial resolutions of multi-task token sequences are gradually increased at different stages, and the attention message at each stage is passed to the next stage to enable cross-scale cross-task interaction. Multi-scale encoder features are utilized to help learn multi-scale information in the InvPT++ decoder. ``$\oplus$'' indicates adding and ``Proj'' is a linear projection layer. 
}
\label{fig:invpt_decoder}
\end{figure}

\begin{figure*}[t]
\centering
\includegraphics[width=1\textwidth]{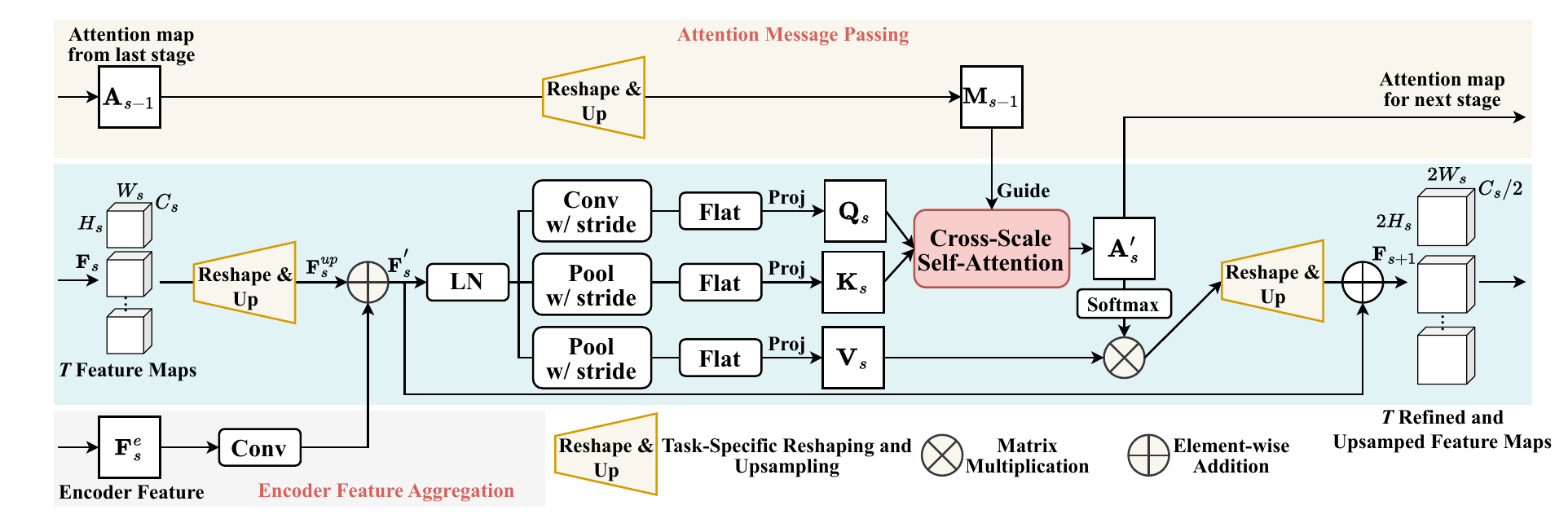}
\caption{
Structure of the designed ``UP-Transformer Block''. It learns to upsample multi-task features, and then establishes global all-task interaction via a novel Cross-Scale Self-Attention module. An Attention Message Passing strategy is designed to bridge attention information between two adjacent stages, and an Encoder Feature Aggregation mechanism helps learn multi-scale information.
The block takes a feature sequence $\mathbf{F}_s$, the attention score matrix from the preceding stage $\mathbf{A}_{s-1}$, and the encoder feature $\mathbf{F}^e_s$ as inputs.
It finally outputs the refined and upsampled multi-task token sequence $\mathbf{F}_{s+1}$ and the attention score matrix $\mathbf{A}_s'$. They are inputs of the next stage.
}
\label{fig:utb}
\end{figure*}

\subsection{Task-Specific Preliminary Decoders}\label{sec:preliminarydecoders}
The function of the task-specific preliminary decoders is to generate task-specific features and coarse predictions for different tasks. 
To accomplish this, we construct a base unit composed of a convolutional layer with $3\times 3$ kernel, a batch normalization operation, and a ReLU, denoted as ``Conv-BN-ReLU". The preliminary decoder of each task consists of two Conv-BN-ReLU units. It accepts the encoder features as input and delivers the task-specific feature for the corresponding task.

Suppose there are $T$ tasks to learn, the preliminary decoders generate task-specific features denoted as $\mathbf{F}_t^d, t \in [1, T]$. The preliminary prediction for the $t$-th task, $\mathbf{P}_t$, is achieved by applying a task-specific $1\times 1$ convolution to the corresponding task-specific feature $\mathbf{F}_t^d$. 
The preliminary predictions are learned with supervision from the ground-truth task labels via loss functions.

Next, we integrate $\mathbf{F}_t^{d}$ and $\mathbf{P}_t$ on the channel dimension. The resulting tensor is processed by a linear projection layer to align the channel dimension to $C_0$, producing the combined task-specific feature $\mathbf{F}_t^{c}$. We then flatten and concatenate all combined task-specific features $\mathbf{F}_t^c, t\in[1,T]$ to form a multi-task token sequence $\mathbf{F}^{c}$, such that $\mathbf{F}^{c} \in \mathbb{R}^{TH_0 W_0 \times C_0}$. This multi-task token sequence, $\mathbf{F}^{c}$, will serve as the input for the subsequent InvPT++ decoder.

\subsection{Structure of InvPT++ Decoder}
\label{sec:invpt_decoder}
The decoder plays a crucial role in InvPT++, facilitating effective collaboration among diverse tasks.
Our design of the InvPT++ transformer decoder building block is motivated by three key considerations:

(i) Establishing interaction among all tasks' features within a spatially global context is crucial for facilitating task synergy in a multi-task model, thus significantly enhancing its performance.

(ii) Many vision transformers~\cite{vit,cvt,pvt} tend to dramatically reduce the resolutions of feature maps due to the prohibitively high computational cost of long-range self-attention at high spatial resolutions. However, the feature map resolution is crucial for the pixel-wise prediction tasks, since the fine-granularity task label value is assigned to each pixel~\cite{pspnet}. Therefore, we want to refine multi-task features at higher resolutions.

(iii) Various tasks necessitate distinct perspectives on visual information. Embracing a multi-scale representation proves immensely advantageous as it offers a holistic understanding of a diverse range of tasks~\cite{mti,hrformer}.

Taking these motivations into account, we design a transformer decoder to progressively enlarge the feature resolution, referred to as the ``InvPT++ decoder''. 
The InvPT++ decoder's architecture is depicted in Fig.~\ref{fig:invpt_decoder}.
The proposed InvPT++ decoder is composed of three stages, each of which features a carefully designed transformer block to facilitate cross-task interaction at varying spatial resolutions.
In the first stage of InvPT++ decoder (\ie, $s=0$), we establish cross-task interaction and refine multi-task features at the encoder's output resolution, \ie, $H_0 \times W_0$. 
During the second and third stages (\ie, $s=1,2$), we design a ``UP-Transformer block'' to progressively increase the feature resolutions and enable cross-task interaction at higher scales. A Cross-Scale Self-Attention module is devised to boost the modeling of cross-task interaction by connecting the attention messages from different scales, and an Encoder Feature Aggregation approach is further developed to assist in learning multi-scale representations.

For clearer illustration in the following, in the $s$-th stage, where $s \in [0,1,2]$, the input feature of  is notated as $\mathbf{F}_s$, and we have $\mathbf{F}_s \in \mathbb{R}^{TH_s W_s \times C_s }$. Here, $H_s$ and $W_s$ denote the height and width of the input feature, while $C_s$ refers to the number of channels.
The input for the initial stage is $\mathbf{F}^c$, as discussed in Sec.~\ref{sec:preliminarydecoders}, leading to $\mathbf{F}_0=\mathbf{F}^c$.

The proposed InvPT++ decoder outputs refined task features for all tasks. It combines the features from various decoder stages to incorporate multi-scale information. 
Specifically, it obtains the token sequences from the outputs at different stages of the InvPT++ decoder and subsequently reshapes them into spatial feature maps.
It aligns the resolution of feature maps from different stages using bilinear interpolation and employs a linear projection layer to match channel numbers before summing the feature maps in an element-wise manner. The merged feature map is then fed into a Conv-BN-ReLU unit to generate the final $T$ refined task features.

\subsection{UP-Transformer Block} 
The UP-Transformer block acts as a key building block in the InvPT++ decoder. It is designed to progressively upsample the multi-task token sequence while fostering cross-task interaction to refine features for all tasks through spatially global long-range interactions.  The structure of the UP-Transformer block is illustrated in Fig.~\ref{fig:utb}. It is used in the second and third stages (\ie, $s=1,2$) of InvPT++ decoder.

\subsubsection{Reshape \& Up: Multi-Task Feature Upsampling} 
\label{sec:reshapeAndUp}
\begin{figure}[t]
\centering
\includegraphics[width=1\linewidth]{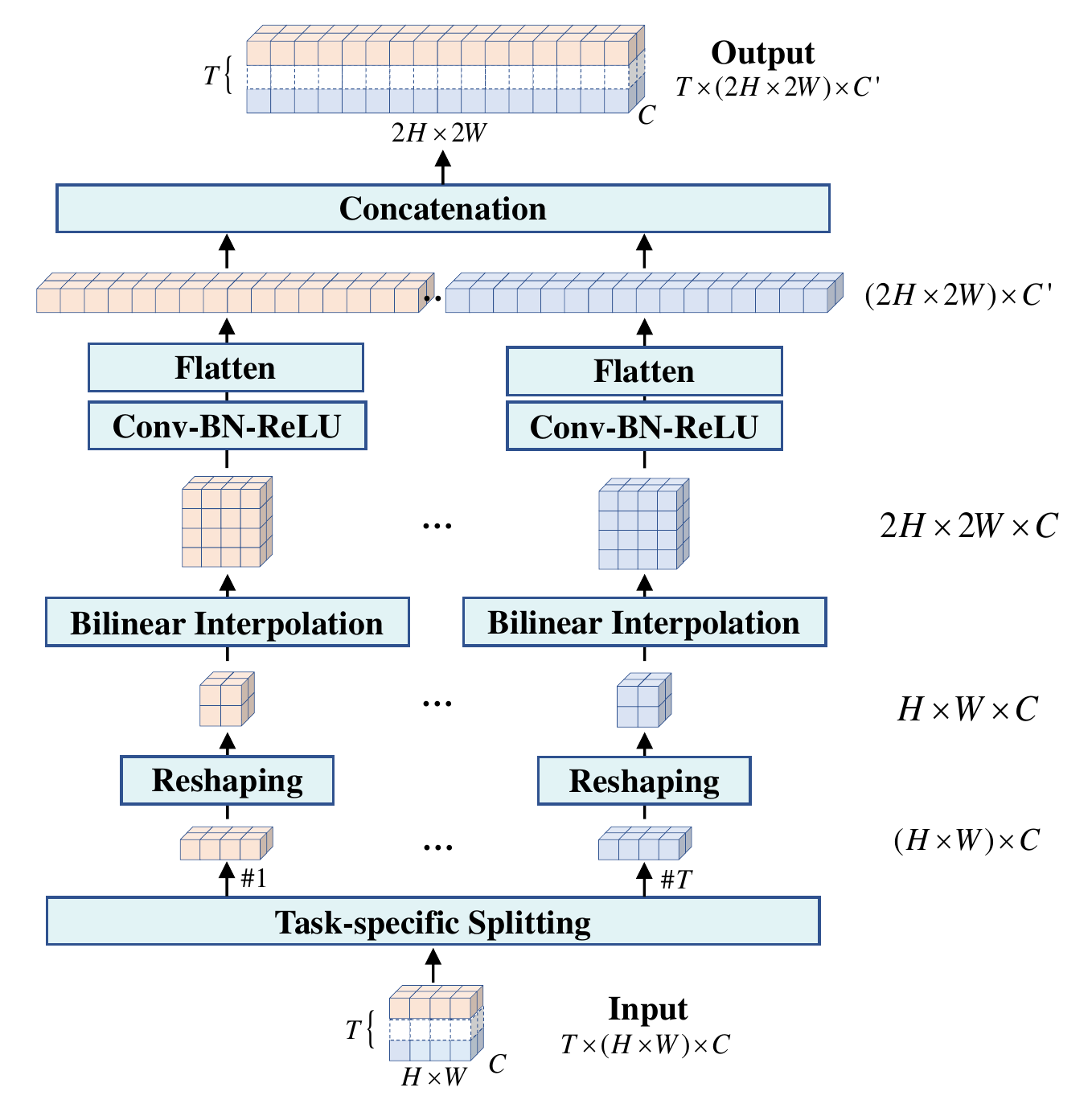}
% \vspace{-10pt}
\caption{Illustration of Reshape \& Up module. The multi-task token sequence is initially divided along the first dimension, creating a series of task-specific token sequences. Each token sequence is then restructured into a spatial feature map and increased in size through bilinear interpolation. Following this, a Conv-BN-ReLU unit is utilized to modify the channel dimension. The final feature maps are then flattened back into token sequences, which are subsequently concatenated to form an upsampled multi-task token sequence.
}
\label{fig:reshapeup}
% \vspace{-18pt}
\end{figure}

To refine task features at higher resolutions, upsampling is considered in the UP-Transformer.
However, directly interpolating the 2D token sequence can disrupt the spatial structure of features, which however is crucial for dense scene understanding problems.
As a result, we design a block to reshape and upsample multi-task token sequences, called the Reshape \& UP module, as depicted in Fig.~\ref{fig:reshapeup}.
In the Reshape \& UP module, we initially divide the token sequence of $T$ tasks $\mathbf{F}_s \in \mathbb{R}^{TH_s W_s \times C_s }$ along the first dimension evenly into $T$ feature groups using tensor slicing.
We then reshape each group's token sequence into a spatial feature map with dimensions $H_s\times W_s\times C_s$.
Each group's feature map is upsampled using bilinear interpolation, doubling the height and width.
The upsampled feature map of each group is further processed by a Conv-BN-ReLU unit to adjust the channel dimension. Finally, the $T$ groups of feature maps are flattened back into token sequences and then concatenated along the first dimension, yielding an upsampled token sequence that contains features for all the $T$ tasks.

At the beginning of UP-Transformer, we employ the Reshape \& Up module  to enhance the spatial resolution of the multi-task token sequence while reducing the channel count by half, yielding an upsampled token sequence $\mathbf{F}^{up}_s \in \mathbb{R}^{4TH_s W_s \times (C_s/2)}$.
Next, $\mathbf{F}_s^{up}$ is combined with a feature token sequence from the transformer encoder, which will be discussed in a later section, and processed by an LN layer\cite{layernorm}. The output, $\mathbf{F}'_s \in \mathbb{R}^{4TH_s W_s \times (C_s/2)}$, is then fed into the subsequent self-attention layer.

\subsubsection{Spatially Global Modeling of Cross-Task Interactions}

In the self-attention block, our goal is to establish spatially global cross-task interaction.  
We utilize the self-attention mechanism to model the relationship between each pair of tokens from the features of all tasks.
However, directly computing the self-attention map based on the high-resolution multi-task token sequence $\mathbf{F}'_s$ would lead to an excessively large memory footprint.
Therefore, we downsample $\mathbf{F}'_s$ prior to inputting it into the self-attention module in all transformer blocks in the InvPT++ decoder.
Specifically, $\mathbf{F}'_s$ is initially split and reshaped into $T$ groups of spatial maps, where each group has a shape of $\mathbb{R}^{2H_s\times 2W_s \times (C_s/2)}$ and corresponds to a single task.
To generate the query tensor, a $3\times 3$ convolution layer with a stride 2 is employed for each group, denoted as $\mathrm{Conv}(\cdot)$.
For generating the key and value tensors, an average pooling operation $\mathrm{Pool}(\cdot, k_s)$ with a stride of $k_s$ ($k_s = 2^{s+1}$ for the $s$-th stage) is utilized.
This approach significantly enhances the computational efficiency of the self-attention module, making spatially global cross-task interactions at high resolutions viable.

Next, the query, key, and value tensors of each group are flattened and concatenated as query, key, and value token sequences.
This process is denoted as $\mathrm{Flatten}(\cdot)$.
Assuming $\mathbf{W}_s^q$, $\mathbf{W}_s^k$, and $\mathbf{W}_s^v$ are the weight matrices of the QKV linear projection layers, the computation of query $\mathbf{Q}_s$, key $\mathbf{K}_s$, and value $\mathbf{V}_s$ can be described as follows:
\begin{equation}
\begin{aligned}
\mathbf{Q}_s &= \mathbf{W}^q_s \times \mathrm{Flatten}\big(\mathrm{Conv}(\mathbf{F}'_s)\big), \, \mathbf{Q}_s \in \mathbb{R}^{{TH_s W_s} \times \frac{C_s}{2}}, \\
\mathbf{K}_s &= \mathbf{W}^k_s \times \mathrm{Flatten}\big(\mathrm{Pool}(\mathbf{F}'_s, k_s)\big), \, \mathbf{K}_s \in \mathbb{R}^{\frac{4TH_sW_s}{(k_s)^2} \times \frac{C_s}{2}},\\
\mathbf{V}_s &= \mathbf{W}^v_s \times \mathrm{Flatten}\big(\mathrm{Pool}(\mathbf{F}'_s, k_s)\big), \, \mathbf{V}_s \in \mathbb{R}^{\frac{4TH_s W_s}{(k_s)^2} \times \frac{C_s}{2}}.
\end{aligned}
\end{equation}

Subsequently, we can calculate the self-attention score matrix $\mathbf{A}_s$ using $\mathbf{Q}_s$ and $\mathbf{K}_s$:
\begin{equation} \label{equ:asco}
    \mathbf{A}_s = \frac{\mathbf{Q}_s \mathbf{K}^T_s}{\sqrt{C'_s}}, \, \mathbf{A}_s \in \mathbb{R}^{T H_s W_s \times \frac{4T H_s W_s}{(k_s)^2}},
\end{equation}
here $C'_s=\frac{C_s}{2}$ serves as a scaling factor to tackle the magnitude explosion issue~\cite{transformer}.
In the original transformer model, the self-attention map is directly computed by applying a softmax activation function to $\mathbf{A}_s$.

As multi-scale information is important for visual understanding tasks~\cite{mti}, we improve the self-attention module by devising a Cross-Scale Self-Attention module that integrates attention information from the preceding layer into the self-attention module at the current layer. 
The subsequent section will discuss the module in detail.

\begin{figure*}[t]
\centering
\includegraphics[width=1\linewidth]{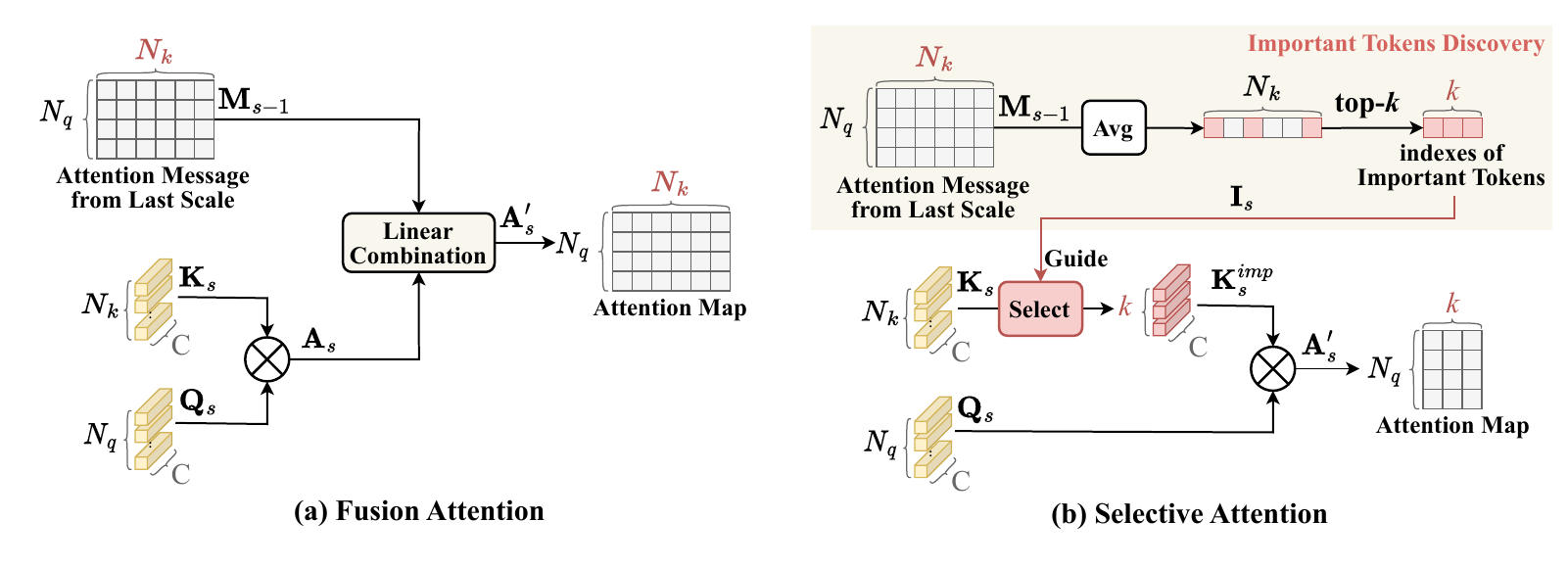}
\vspace{-25pt}
\caption{Illustration of two types of Cross-Scale Self-Attention modules: Fusion Attention and Selective Attention. \textbf{(a)} Fusion Attention merges cross-scale attention messages from the preceding layer with the attention map in the present stage. This facilitates the learning process of self-attention modules across different layers. \textbf{(b)} In contrast, Selective Attention suggests using the attention message from the earlier scale as guidance to pinpoint important tokens in self-attention computation, which significantly cuts down computational redundancy. It chooses the top-$k$ tokens having the highest average values in the attention message and incorporates only these $k$ tokens in the current stage's self-attention calculation.}
\label{fig:cross_scale_self_attn}
% \vspace{-18pt}
\end{figure*}

\subsection{Cross-Scale Self-Attention: Fusion Attention and Selective Attention} 
As shown in Fig.~\ref{fig:utb}, to facilitate cross-task interaction modeling of multi-scale features across different scales, we propose a special self-attention module named Cross-Scale Self-Attention in the UP-Transformer block. The Cross-Scale Self-Attention mechanism passes the attention messages from the previous stage to the current stage to help the learning of cross-task interaction with cross-scale information.

To obtain an attention message from the previous stage, we first design an Attention Message Passing mechanism.
As shown in Fig.~\ref{fig:utb}, the Attention Message Passing conveys the attention score matrix from the $(s-1)$-th stage, $\mathbf{A}_{s-1} \in \mathbb{R}^{\frac{T H_s W_s}{4} \times \frac{4T H_s W_s}{(k_s)^2}}$, to the current stage. 
Specifically, we first properly adjust the stride parameters of the average pooling layers $k_s$ in order to keep the second dimension of $\mathbf{A}_s$ consistent across various stages. 
The Reshape \& Up module is utilized to increase the spatial resolution of  $\mathbf{A}_{s-1}$ along the first dimension, resulting in an attention message matrix $\mathbf{M}_{s-1} \in \mathbb{R}^{{T H_s W_s} \times \frac{4T H_s W_s}{(k_s)^2}}$ with dimensions that match those of $\mathbf{A}_s$.
The attention message matrix $\mathbf{M}_{s-1}$, which embodies the cross-task relationship from the previous stage, will be employed to assist in modeling cross-task interaction in the current phase through the implementation of Cross-Scale Self-Attention as discussed further on.

We develop two different types of Cross-Scale Self-Attention mechanisms, \emph{i.e.}, Fusion Attention and Selective Attention, as shown in Fig.~\ref{fig:cross_scale_self_attn}. These two mechanisms vary in their utilization of the attention message from the preceding layer. In essence, Fusion Attention employs a linear combination to merge the attention message from the previous stage and the attention score at current stage, whereas the Selective Attention introduces a more efficient and effective strategy for learning cross-task interactions guided by attention message from the previous stage. We will introduce these two variants in the following sections.

\renewcommand{\arraystretch}{1.7}
\begin{table}[t]
\setlength{\tabcolsep}{4pt}
% \vspace{-20pt}
\centering
\caption{Details about the shapes of query $\mathbf{Q}$, key $\mathbf{K}$, and value $\mathbf{V}$ tensors in different stages of InvPT++ decoder. }
\label{tab:qkv}
\resizebox{0.8\linewidth}{!}{
    \begin{tabular}{cccc}
    \hline
  & $s=0$   & $s=1$  & $s=2$   \\
    \hline
     $\mathbf{Q}$  & $\frac{TH_0 W_0}{4}, C_0$  & $TH_0 W_0, \frac{C_0}{2}$ &  $4TH_0 W_0, \frac{C_0}{4}$\\
     \hline
     $\mathbf{K}$ & $\frac{TH_0 W_0}{4}, C_0$ &  $\frac{TH_0 W_0}{4}, \frac{C_0}{2}$ & $\frac{TH_0 W_0}{4}, \frac{C_0}{4}$\\
      \hline
     $\mathbf{V}$ & $\frac{TH_0 W_0}{4}, C_0$  &  $\frac{TH_0 W_0}{4}, \frac{C_0}{2}$  & $\frac{TH_0 W_0}{4}, \frac{C_0}{4}$\\
     \hline
    \end{tabular}}
% \vspace{-2pt}
\end{table}
\renewcommand{\arraystretch}{1}

\subsubsection{Fusion Attention}
Fusion Attention learns a pair of weights for fusing the attention message $\mathbf{M}_{s-1}$ from the previous stage and the attention score $\mathbf{A}_s$ at the current stage, as shown in Fig.~\ref{fig:cross_scale_self_attn}. 
Specifically, after we obtain the attention message matrix $\mathbf{M}_{s-1}$, Fusion Attention 
learns two weights $\alpha^1_s$ and $\alpha^2_s$ to combine $\mathbf{A}_s$ and $\mathbf{M}_{s-1}$, and get the combined attention map $\mathbf{A}_s^m$ after applying a Softmax function:
\begin{equation}
\begin{aligned}
\mathbf{A}'_{s} &=  \alpha^1_s \mathbf{A}_{s} +  \alpha^2_s \mathbf{M}_{s-1},\\
\mathbf{A}_s^{m} &= \mathrm{Softmax} (\mathbf{A}'_s),
\end{aligned}
\end{equation}
here the $\mathrm{Softmax(\cdot)}$ function applies a row-wise softmax operation to the composite attention score matrix.

The UP-Transformer block outputs the updated multi-task token sequence $\mathbf{F}_{s+1}$, which is derived by multiplying  $\mathbf{A}^{m}_s$ with the value tensor $\mathbf{V}_s$. 
Subsequently, $\mathbf{F}_{s+1}$  undergoes upsampling and is added by the skip connection originating from  $\mathbf{F}'_s$. 

\subsubsection{Selective Attention}
Selective Attention takes a different approach from Fusion Attention and achieves much higher computational efficiency.
Its motivation is that, since the attention message matrix $\mathbf{M}_{s-1}$ encompasses the token-level cross-task relationships acquired from the preceding layer, we can leverage this information to find the important tokens and performs self-attention exclusively among them. This approach reduces computational costs by focusing only on the essential tokens. Consequently, Selective Attention reduces redundant token relationships in the attention map, enhancing the effectiveness and efficiency of cross-task interaction.

As illustrated in Fig.~\ref{fig:cross_scale_self_attn}, we first carry out a process named ``important tokens discovery'' to pinpoint the important tokens. 
Specifically, suppose the attention message matrix $\mathbf{M}_{s-1}$ has a dimension of $N_q \times N_k$, we first compute the average value of $\mathbf{M}_{s-1}$ along the first dimension and get $N_k$ average values, we denote this process as ``$\mathrm{Avg}(\cdot)$''. These average values represent the significance levels of the corresponding key tokens in the self-attention of the preceding scale. Subsequently, the top-$k$ key tokens are selected as important tokens based highest average values, and their corresponding indexes are recorded as $\mathbf{I}_s \in \mathbb{R}^k$. We denote this operation as ``$\mathrm{top}(·, k)$", leading to the following:
\begin{equation}
\begin{aligned}
\mathbf{I}_s = \mathrm{top}(\mathrm{Avg}(\mathbf{M}_{s-1}), k). 
\end{aligned}
\end{equation}

\begin{figure*}[!t]
	\centering
	\includegraphics[width=1\textwidth]{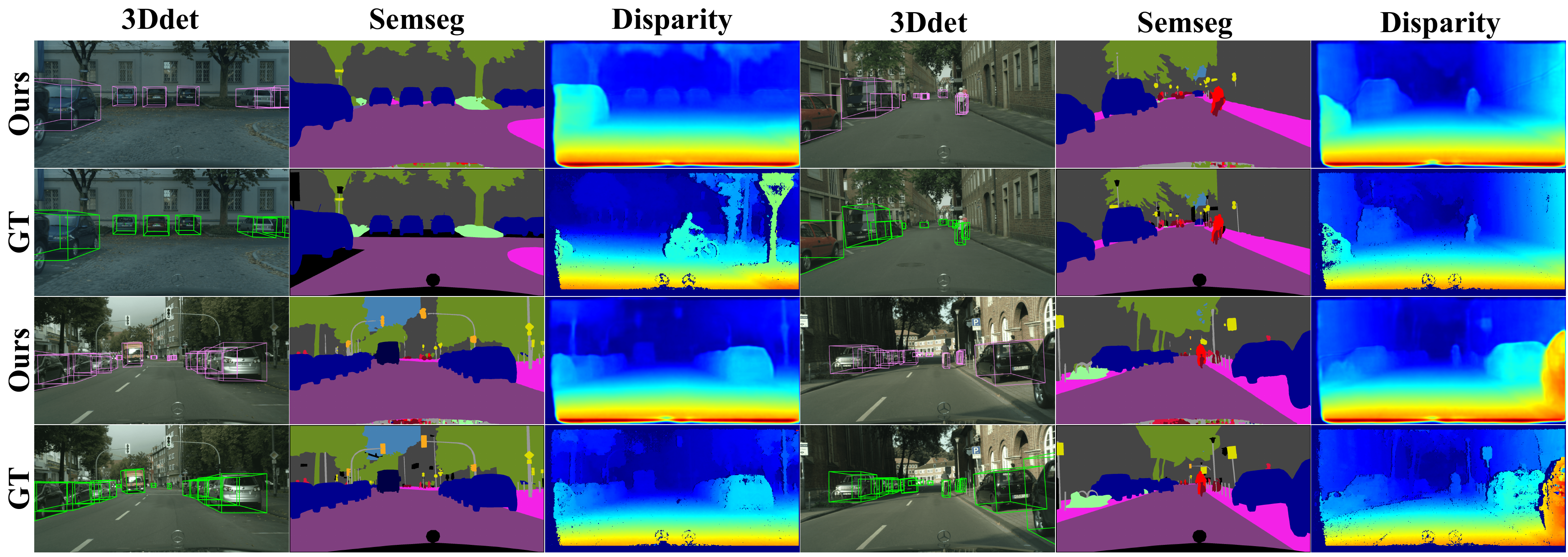}
	\caption{Qualitative comparison between our predictions and the ground-truth labels (GT) for joint 2D-3D multi-task scene understanding using the Cityscapes-3D dataset\cite{gahlert2020cityscapes3d}.
	}
\label{fig:vis_cityscapes}
\end{figure*}

We then select the important key tokens of current stage $\textbf{K}_s^{imp}$ from $\textbf{K}_s$ based on the indexes $\mathbf{I}_s$. This operation is notated as ``$\mathrm{Select}(\textbf{K}_s, \mathbf{I}_s)$''.
Subsequently, we compute the attention map exclusively for the $k$ important key tokens $\textbf{K}_s^{imp}$, which prevents redundant computation of all tokens, thereby enhancing efficiency and effectiveness. The computation of attention map $\mathbf{A}_s^{m}$ is now formulated as:
\begin{equation}
\begin{aligned}
\textbf{K}_s^{imp} &= \mathrm{Select}(\textbf{K}_s, \mathbf{I}_s),\\
\mathbf{A}'_{s} &= 
\frac{\mathbf{Q}_s {\mathbf{K}^{imp}_s}^T}{\sqrt{C'_s}},\\
\mathbf{A}_s^{m} &= \mathrm{Softmax} (\mathbf{A}'_s).
\end{aligned}
\end{equation}
In InvPT++, we favor Selective Attention as the standard Cross-Scale Self-Attention method, due to its superior efficiency and effectiveness. These merits are further substantiated in the experimental section.

\subsubsection{Encoder Feature Aggregation}
Different scene understanding tasks depend on distinct levels of visual information. Some tasks, such as object boundary detection, necessitate a basic semantic comprehension of images, while others, such as semantic segmentation and human parsing, demand higher-level visual representation. 
Consequently, learning features at different scales plays a crucial role in providing comprehensive information for multi-task prediction. With this motivation, we introduce an effective Encoder Feature Aggregation approach that integrates multi-scale features from the encoder into corresponding decoder stages.
As depicted in Fig.~\ref{fig:utb}, within each UP-Transformer block, we obtain the corresponding encoder token sequence with the same spatial resolution, denoted as $\mathbf{F}^e_s$. Assuming the channel count of $\mathbf{F}^e_s$ is $C_s^e$, we reshape it into a spatial feature map and employ a $3\times 3$ convolution layer to modify the channel count to $C_s$. Subsequently, we flatten the result back into a token sequence and expand it by $T$ times to match the dimensions of $\mathbf{F}_s^{up}$ before adding to $\mathbf{F}_s^{up}$.
In doing so, the InvPT++ decoder receives multi-scale features from the encoder, which ultimately enhances the overall performance.

\begin{figure*}[!t]
	\centering
	\includegraphics[width=1\textwidth]{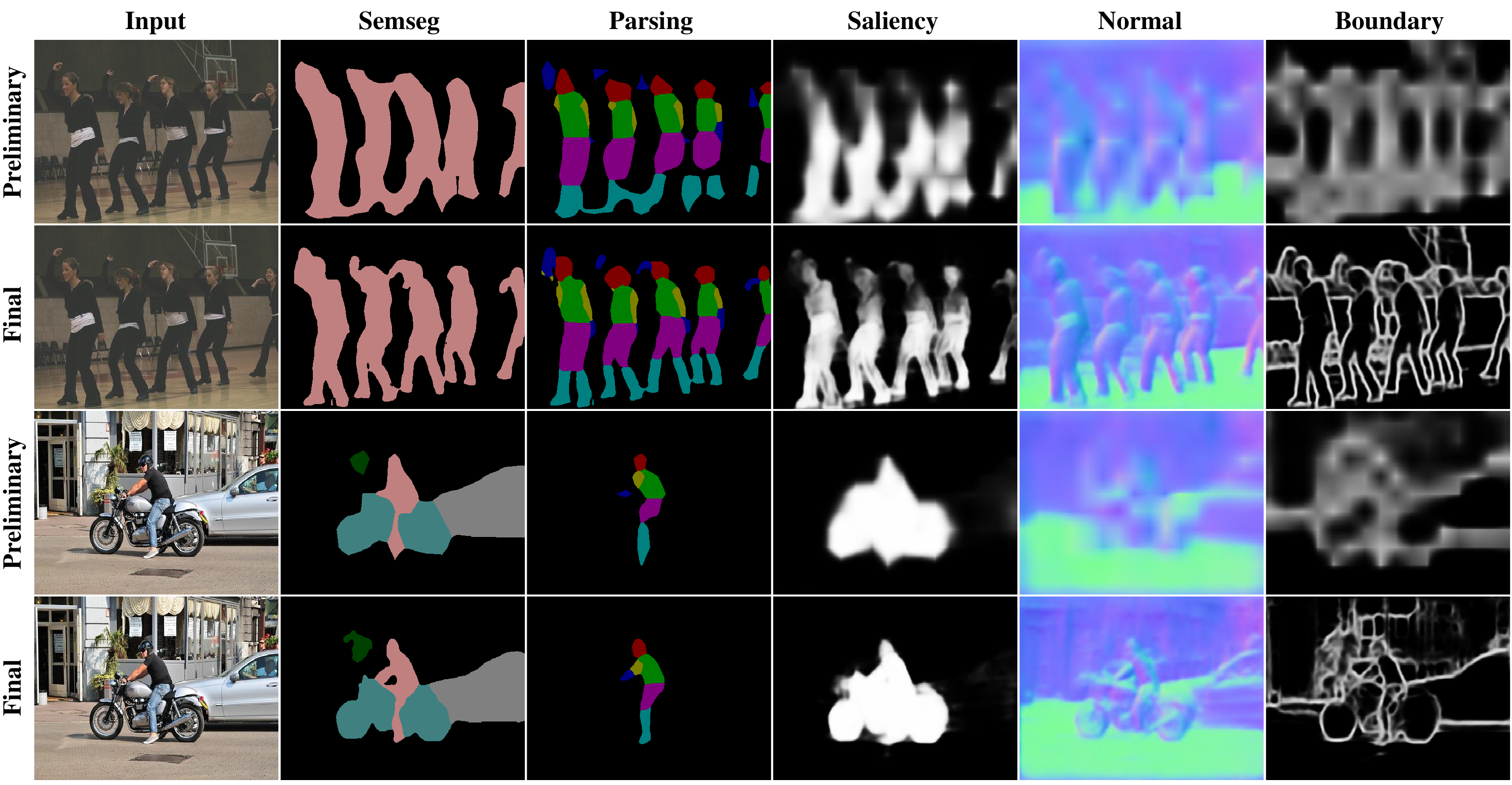}
	\caption{Qualitative analysis of the preliminary decoder's predictions and the final predictions of the InvPT++ decoder on PASCAL-Context. The final predictions are notably more accurate across all tasks.
	}
	% \vspace{-15pt}
	\label{fig:vis_inter}
\end{figure*}

\section{Experiments}
\label{sec:experiments}
In the experimental section, our goal is to present comprehensive experimental findings that evaluate the efficacy of our proposed approach.

\subsection{Implementation of Experiments}
\subsubsection{Datasets} 
To assess our approach, we use four widely recognized benchmarks in multi-task learning which include PASCAL-Context~\cite{chen2014detect,everingham2010pascal}, NYUD-v2~\cite{silberman2012indoor}, Cityscapes~\cite{Cordts2016Cityscapes}, and Cityscapes-3D~\cite{gahlert2020cityscapes3d} datasets.
\textbf{PASCAL-Context} dataset consists of 4,998 training samples and 5,105 testing samples, featuring labels for various essential visual understanding tasks, \emph{e.g.}, semantic segmentation (Semseg), human parsing (Parsing), and object boundary detection (Boundary). Earlier work in multi-task learning generates pseudo labels for two more tasks~\cite{astmt}: surface normal estimation (Normal) and saliency detection (Saliency).
\textbf{NYUD-v2} dataset contains 1,449 training samples and 795 testing samples, offering labels for Semseg and Depth. Normal and Boundary labels can be derived from these annotations. We evaluate our approach using all tasks available in these datasets.
\textbf{Cityscapes} dataset includes images of streets across different cities, providing pixel labels for Semseg and Depth. This dataset consists of 2,975 training images and 500 validation images with fine annotations.
\textbf{Cityscapes-3D} dataset further supplies 3D bounding boxes for various vehicles in Cityscapes images, allowing us to perform joint learning of 3D object detection (3Ddet), Semseg, and Depth.

\subsubsection{Evaluation}
We employ the same set of evaluation metrics as utilized in previous works~\cite{mti,invpt2022}, including, 
the mean Intersection over Union (mIoU) for Semseg and Parsing tasks, the Root Mean Square Error (RMSE) or the absolute error (absErr) for the Depth task, the mean error (mErr) of angles for the Normal task, the maximal F-measure (maxF) for the Saliency task, and the optimal-dataset-scale F-measure~(odsF) for the Boundary task.
The 3Ddet task is evaluated by mean detection score~(mDS)~\cite{gahlert2020cityscapes3d}.
We compute the mean relative difference across all tasks in comparison to the single-task baseline, denoted as MT Gain $\Delta_m$~\cite{astmt}.

\begin{figure*}[t]
\centering
\begin{subfigure}[t]{\textwidth}
    \centering
    \includegraphics[width=\textwidth]{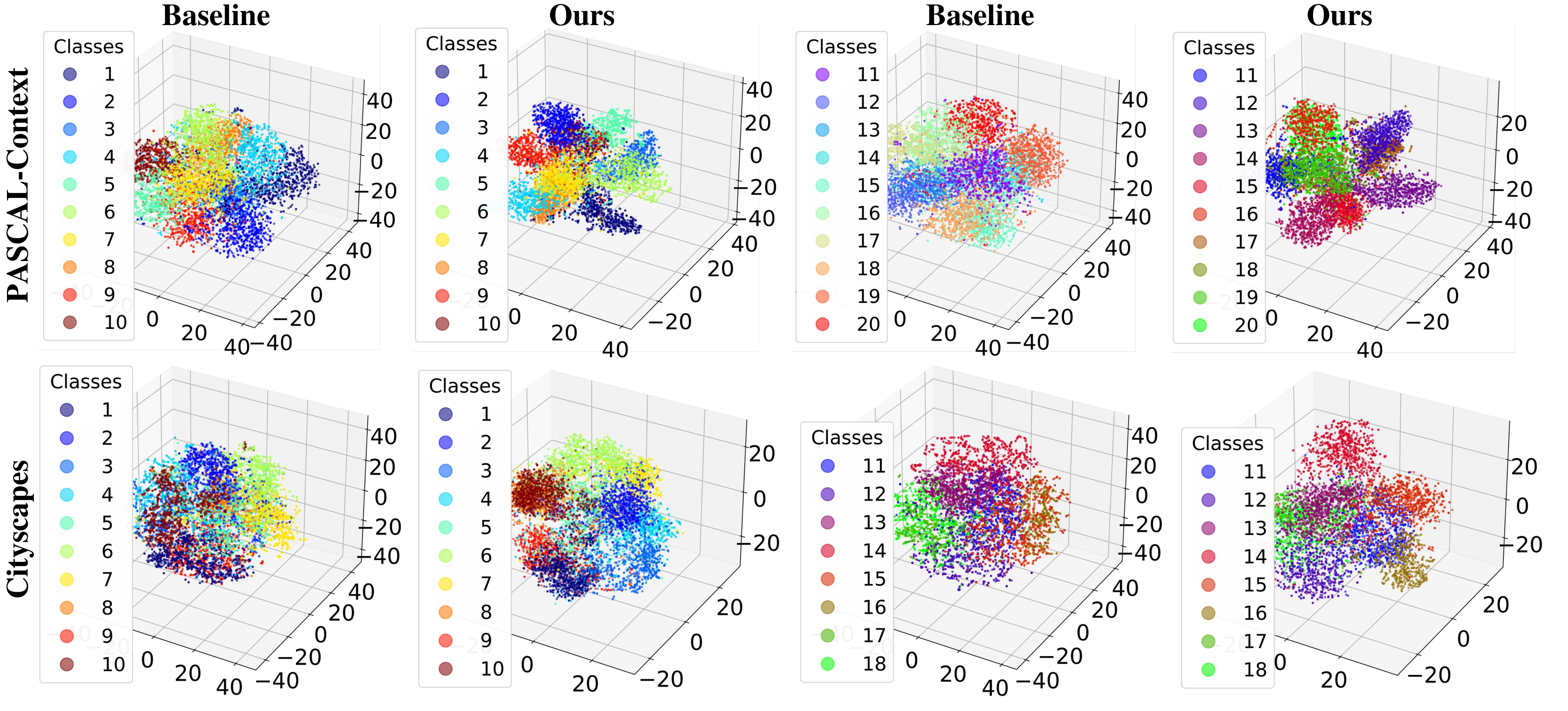}
    \caption{Visualization of learned features with t-SNE~\cite{van2008visualizing} of all 20 semantic classes on PASCAL-Context and Cityscapes datasets. }
    \label{fig:tsne}
\end{subfigure}

\begin{subfigure}[t]{\textwidth}
    \centering
    \includegraphics[width=\textwidth]{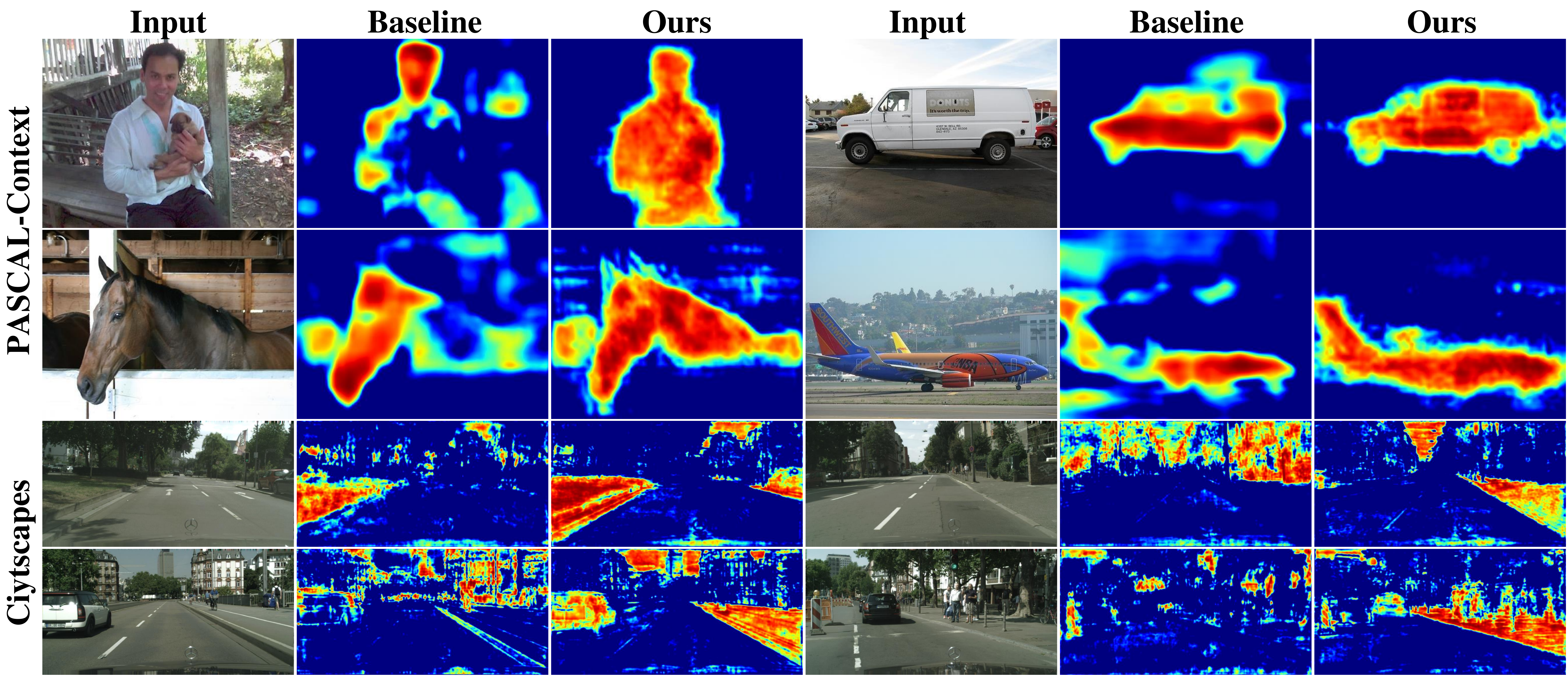}
    \caption{Some examples of learned features for the semantic segmentation task on PASCAL-Context and Cityscapes datasets.}
    \label{fig:featimprove}
\end{subfigure}
\caption{Qualitative study of the learned features from the baseline model and our model. We find that InvPT++ generates much more discriminative features than the baseline model.}
\label{fig:tsne_featimprove}
\vspace{-18pt}
\end{figure*}

\subsubsection{Implementation Details}
For model training, we follow the setting of InvPT~\cite{invpt2022} and use Adam optimizer to train the models. The number of training iterations and the batch size are 40,000 and 4, respectively. We adopt the Polynomial learning rate scheduler, using a learning rate of $2\times 10^{-5}$ and a weight decay rate of $1\times 10^{-6}$. 
We adopt the design of loss functions for different tasks in MTI-Net~\cite{mti} and use the same loss weights. 
The Swin-T transformer~\cite{swin} pre-trained on ImageNet-22K~\cite{deng2009imagenet} is considered as the encoder for our ablation study. 
The preliminary decoders have an output channel count of 768.
In experiments on Cityscapes, we employ ViT-B~\cite{vit} as the encoder and train the models for 80,000 iterations with a batch size of 8. For experiments on Cityscapes-3D, due to the higher resolution requirement of $768\times 1536$ for 3D detection, the Swin-T is used as the encoder and it is trained with 40,000 iterations using a batch size of 2.
For 3Ddet, we employ the final prediction heads of FCOS-3D~\cite{wang2021fcos3d} to estimate location coordinates, rotation angles, dimensions, object classes, center-ness, and direction classification. 
For a comprehensive understanding of the 3D detection prediction heads, please see the FCOS-3D paper for more details.

\begin{figure}[!t]
\centering
\includegraphics[width=\linewidth]{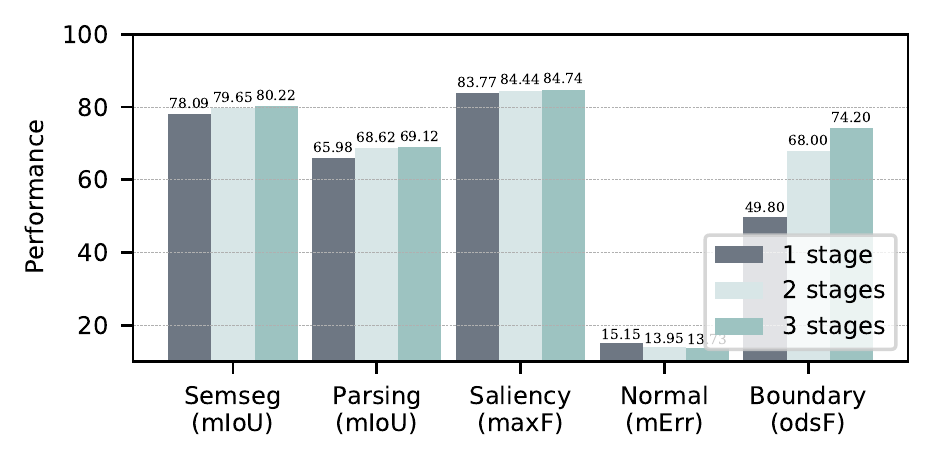}
\caption{Ablation study of different numbers of stages in the InvPT++ decoder. More stages lead to better performance on all tasks in InvPT++.}
\label{fig:stages}
\end{figure}

\subsubsection{Implementation of Encoder Feature Aggregation}

The design of our Encoder Feature Aggregation modules is tailored to the distinct attributes of various encoder architectures. For encoder models with clearly defined hierarchical structures, such as the Swin Transformer~\cite{swin}, we can directly feed their encoder token sequences from different stages into the InvPT++ decoder.

In contrast, for models that exhibit a uniform structure across all layers, such as ViT~\cite{vit}, we adopt a different approach. Here, we manually choose token sequences from different layers in a uniform distribution. These selected token sequences are then reshaped into spatial maps and passed through transposed convolution layers. This operation aligns the spatial resolution of these sequences with the decoder features at the target stage.

Specifically, in the case of the ViT-B model, which comprises 12 layers in total, we opt to use the features from the third, sixth, and ninth layers of the encoder. For the ViT-L model, which has 24 layers, we utilize the features from the sixth, twelfth, and eighteenth layers of the encoder.
For the feature map of the first scale, we employ a transposed convolution with a kernel size of 2 and a stride of 2. For the feature map of the second scale, we use a kernel size of 4 and a stride of 4.

\subsubsection{Details about QKV in InvPT++ Decoder} 
As introduced in the method section, the query $\mathbf{Q}$, key $\mathbf{K}$, and value $\mathbf{V}$ tensors have different shapes at different stages of InvPT++ decoder. We provide more details about their shapes in Table~\ref{tab:qkv}. The notations are defined in Section~\ref{sec:method}.

\subsubsection{Model Optimization} 
For experiments on NYUD-v2 and PASCAL-Context datasets, we take into account a total of six dense prediction tasks, \emph{i.e.}, Segseg, Depth, Normal, Parsing, Saliency, and Boundary.
For continuous regression tasks (\ie, Depth and Normal), an $\mathcal{L}1$ loss is employed. For discrete classification tasks (\ie, Semseg, Parsing, Saliency, and Boundary), a cross-entropy loss is utilized. To maintain simplicity, we apply the same set of loss functions for both intermediate and final supervision. The entire model can be optimized end-to-end.
In experiments on Cityscapes, we employ cross-entropy loss for Semseg and $\mathcal{L}1$ loss for Depth. 
For 3D object detection, we utilize a combination of loss functions as described in the FCOS-3D~\cite{wang2021fcos3d}.

\subsubsection{Data Pre-Processing} 
We adopt the same data pre-processing method as \cite{atrc}.
On PASCAL-Context, images are padded to a size of $512\times 512$, while on NYUD-v2, input images are randomly cropped to $448\times 576$, since Swin Transformer~\cite{swin} requires even height and width dimensions for patch merging. Common data augmentations, such as random scaling, cropping, horizontal flipping, and color jittering, are employed.
For experiments on Cityscapes, we adhere to the data pre-processing of UR~\cite{li2022universal}. We use the 7-class Semseg annotation and set the image resolution to $128\times 256$. Random flip and random scale data augmentation methods are utilized.
For experiments on Cityscapes-3D, we use 19-class Semseg annotation and set the image resolution to $768\times 1536$. Data augmentation is not applied to Cityscapes-3D.

\subsection{Model Analysis}
\subsubsection{Baselines and Model Variants} 
We establish a series of model baselines and variants to validate different components of the proposed InvPT++, as presented in Table~\ref{tab:abl_modules_nyudpascal}.
The model variants are illustrated as follows:
(i) InvPT++ Baseline (MT) represents a robust multi-task model. It employs Swin-T as the transformer encoder and utilizes 2 Conv-BN-ReLU units with a $3\times 3$ kernel size as the task-specific decoder for each task, which is the same as the preliminary decoder used in our method. The encoder feature map is upsampled 8 times before the decoders for finer-granularity predictions, using a multi-scale feature aggregation strategy to further enhance performance, similar to multi-task baselines in previous works~\cite{mti,atrc}.
(ii) InvPT++ Baseline (ST) is a collection of single-task models with a structure akin to InvPT++ Baseline (MT).
(iii) InvPT++ w/ UTB incorporates the UP-Transformer block into InvPT++ Baseline (MT).
(iv) InvPT++ w/ UTB + FA further implements Attention Message Passing with Fusion Attention (FA).
(v) InvPT++ w/ UTB + FA + EFA introduces Encoder Feature Aggregation.

\begin{table}[!t]
\centering
\caption{Comparison of Fusion Attention (Fusion) and Selective Attention (Selective) on PASCAL-Context. Selective Attention attains superior performance across all tasks while requiring significantly less computational resources.}
\label{tab:sa_fa_pascal}
\huge
\resizebox{1.\linewidth}{!}{
    \begin{tabular}{ccccccc}
    \toprule
    \multirow{2}*{ \textbf{Version} }  & \textbf{Semseg}  & \textbf{Parsing}
       & \textbf{Saliency} & \textbf{Normal} & \textbf{Boundary} & \multirow{1}*{\textbf{Decoder}}
     \\
        & mIoU $\mathbf{\uparrow}$  & mIoU $\mathbf{\uparrow}$
       & maxF $\mathbf{\uparrow}$ & mErr $\mathbf{\downarrow}$ & odsF $\mathbf{\uparrow}$& \multirow{1}*{\textbf{FLOPs}}
     \\
    \midrule
    Fusion & {79.03} & {67.61} & {84.81} & {14.15} & {73.00} & -\\
    Selective & \textbf{80.22} & \textbf{69.12} & \textbf{84.74} & \textbf{13.73} & \textbf{74.20} & \textbf{-22.51\%}\\
    \bottomrule
    \end{tabular}}
\end{table}

\begin{table}[!t]
\centering
\caption{Comparison of Fusion Attention (Fusion) and Selective Attention (Selective)  on NYUD-v2. Selective Attention achieves higher or equal performance on all tasks compared with Fusion Attention.}
\label{tab:sa_fa_nyud}
\resizebox{.86\linewidth}{!}{
    \begin{tabular}{ccccccc}
    \toprule
    \multirow{2}*{ \textbf{Version} }   & \textbf{Semseg}  & \textbf{Depth}  & \textbf{Normal} & \textbf{Boundary}  \\
         &mIoU $\mathbf{\uparrow}$ & RMSE $\mathbf{\downarrow}$ & mErr $\mathbf{\downarrow}$ & odsF $\mathbf{\uparrow}$ \\
    \midrule
     Fusion &53.56 & 0.5183 & 19.04 & \textbf{78.10}\\
     Selective & \textbf{53.85} & \textbf{0.5096} & \textbf{18.67} & \textbf{78.10}\\
    \bottomrule
    \end{tabular}}
\end{table}

\begin{figure}[!t]
	\centering
	% \vspace{-5pt}
	\includegraphics[width=1\linewidth]{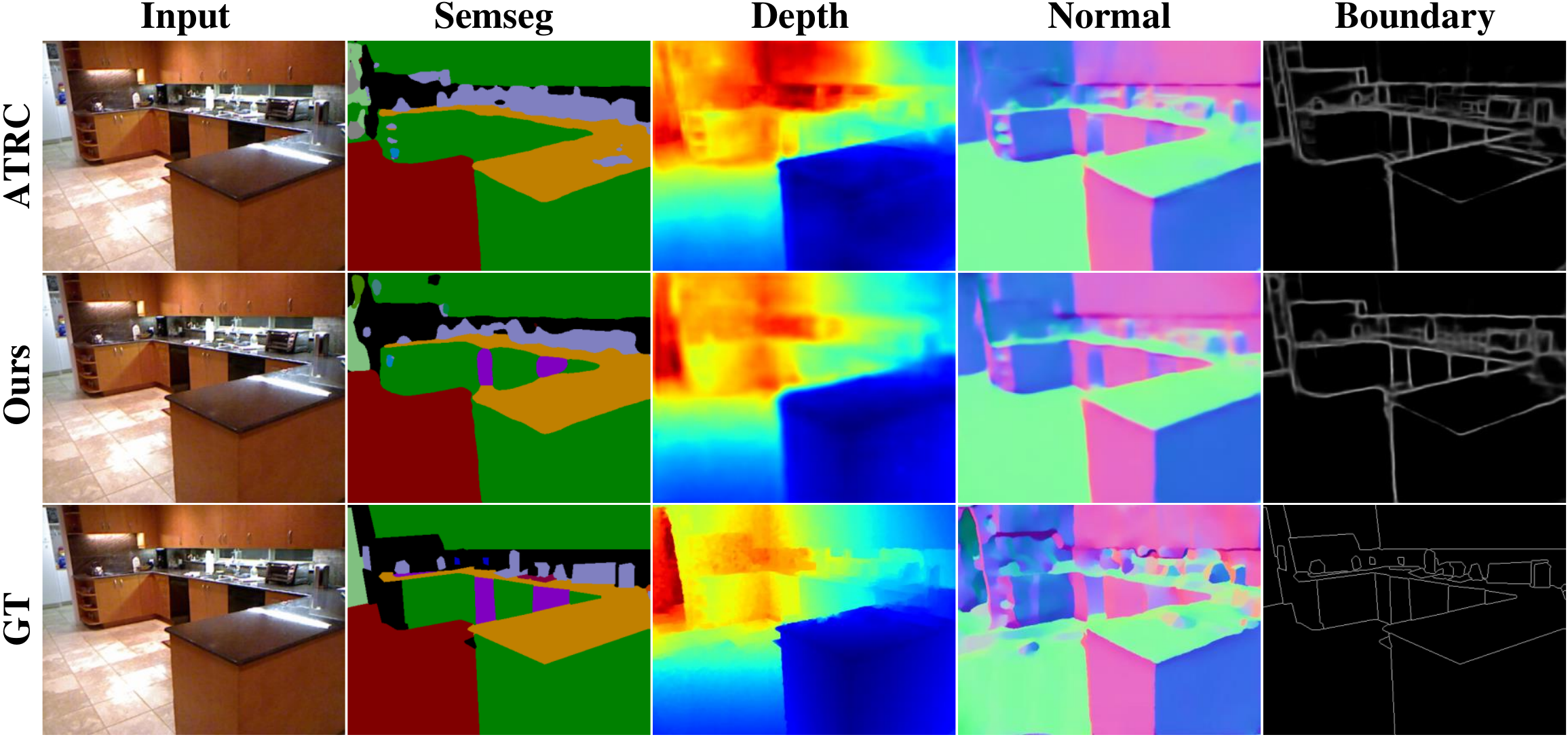}
	\caption{Qualitative analysis of the predicted results on NYUD-v2. Our InvPT++ generates predictions with higher quality on all tasks.
	}
	% \vspace{-20pt}
	\label{fig:qualitative_sota_nyud}
\end{figure}

\begin{figure*}[!t]
\centering
\includegraphics[width=\textwidth]{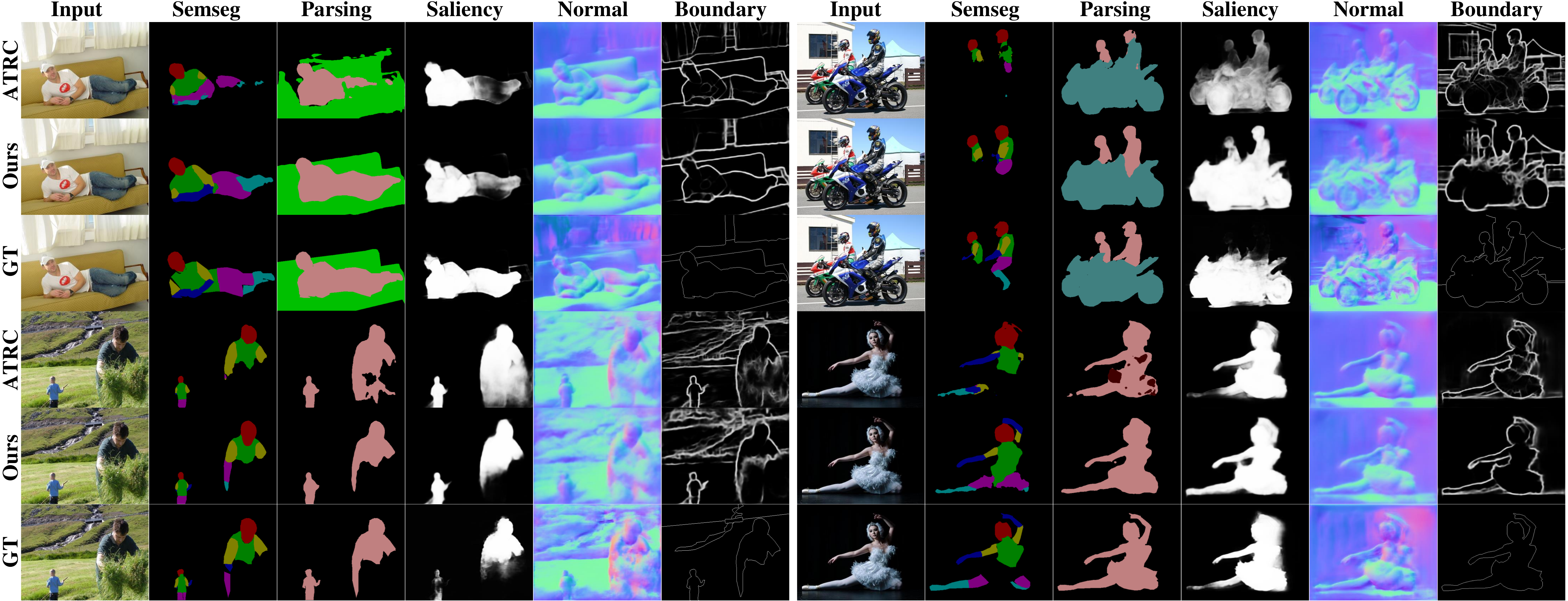}
\caption{
Visualization of the generated multi-task predictions by ATRC~\cite{atrc} and InvPT++. InvPT++ produces notably improved predictions, particularly in higher-level tasks including semantic segmentation and human parsing.
}
\label{fig:qualitative_pascal}
\end{figure*}

\begin{table*}[t]
\centering
\caption{Ablation study of the proposed modules in InvPT++ decoder. Our proposals demonstrate consistent improvement across various datasets and deliver a clear overall enhancement for each individual task and multi-task (MT) gain $\Delta_m$. A Swin-T encoder is employed in all experiments. '$\mathbf{\downarrow}$' signifies that lower values are better, while '$\mathbf{\uparrow}$' indicates higher values are better.}
\small
\label{tab:abl_modules_nyudpascal}
\resizebox{1.\linewidth}{!}{
    \begin{tabular}{l|cccc|cccccccc}
    \toprule
        \multicolumn{1}{c|}{ \multirow{3}*{ \textbf{Model}} }   & \multicolumn{4}{c|}{\textbf{NYUD-v2}} & \multicolumn{5}{c}{\textbf{PASCAL-Context}} \\
        \cline{2-10}
        & \textbf{Semseg}  & \textbf{Depth}  & \textbf{Normal} & \textbf{Boundary} & \textbf{Semseg}  & \textbf{Parsing}  & \textbf{Saliency} & \textbf{Normal} & \textbf{Boundary}  \\
        &mIoU $\mathbf{\uparrow}$ & RMSE $\mathbf{\downarrow}$ & mErr $\mathbf{\downarrow}$ & odsF $\mathbf{\uparrow}$ & mIoU $\mathbf{\uparrow}$  & mIoU $\mathbf{\uparrow}$
       & maxF $\mathbf{\uparrow}$ & mErr $\mathbf{\downarrow}$ & odsF $\mathbf{\uparrow}$ \\
    \midrule
    InvPT++ Baseline (ST)& 43.29& 0.5975 & 20.80 & 76.10 & 72.43 & 61.13 & 83.43 & 14.38 &  71.50 \\
     \midrule
    InvPT++ Baseline (MT) & 41.06  & 0.6350 & 21.47& 76.00 & 70.92 & 59.63 & 82.63 & 14.63 & 71.30\\
    InvPT++ w/ UTB  & 43.18 & 0.5643 & 21.05 & 76.10 & 72.34 & 61.08 & 83.99 & 14.49 & 71.60  \\
    InvPT++ w/ UTB+FA &  43.64 & 0.5617 & 20.87 & 76.10 & 73.29 & 61.78 & 84.03 & 14.37 & 71.80 \\
     InvPT++ w/ UTB+FA+EFA & 44.27 & 0.5589 & 20.46 & 76.10 & 73.93 & 62.73 & 84.24 & 14.15& 72.60\\
     \midrule
     {MT Gain $\Delta_m$}~\cite{astmt}& \multicolumn{4}{c|}{+\textbf{2.59}} & \multicolumn{5}{c}{+\textbf{1.76}} \\
    \bottomrule
    \end{tabular}}
\end{table*}

\begin{table}[t]
\centering
\caption{Experiments of using different ratios of selected tokens in Selective Attention on PASCAL-Context dataset. Selecting more tokens tends to bring better performance.}
\label{tab:token_ratios}
\resizebox{1.\linewidth}{!}{
    \begin{tabular}{ccccccc}
    \toprule
    \textbf{Retention}  & \textbf{Semseg}  & \textbf{Parsing}
       & \textbf{Saliency} & \textbf{Normal} & \textbf{Boundary}
     \\
        \textbf{Ratio}& mIoU $\mathbf{\uparrow}$  & mIoU $\mathbf{\uparrow}$
       & maxF $\mathbf{\uparrow}$ & mErr $\mathbf{\downarrow}$ & odsF $\mathbf{\uparrow}$
     \\
    \midrule
    25\% & 80.01 & 69.06 & 84.71 & \textbf{13.71} & 74.10 \\
    50\%  & \textbf{80.22} & {69.12} & {84.74} & {13.73} & {74.20} \\
    75\% & 79.99 & \textbf{69.27} & \textbf{84.92} & \textbf{13.71} & \textbf{74.30}\\
    \bottomrule
    \end{tabular}}
\end{table}

\subsubsection{Ablation Study of InvPT++ Decoder}
We conduct a series of experiments on PASCAL-Context and NYUD-v2 to validate the efficacy of different modules of InvPT++, especially in the decoder, and show the results in Table~\ref{tab:abl_modules_nyudpascal}. Based on the experimental results, we observe that the proposed UP-Transformer Block (UTB), which is a core contribution, brings significant performance improvement on Semseg by 2.12, on Depth by 0.0707 and on Normal by 0.42, on the NYUD-v2 dataset. 
Introducing the Fusion Attention (FA) version of Cross-Scale Self-Attention results in notable performance improvement, which is further boosted by applying the proposed Encoder Feature Aggregation~(EFA).

To benchmark against single-task baselines, we compare our model with its single-task equivalent, termed "InvPT Baseline (ST)", on both datasets as displayed in Table~\ref{tab:abl_modules_nyudpascal}.  Our model exhibits a substantial performance elevation compared with the single-task variant, achieving a multi-task gain of $\textbf{2.59\%}$ on NYUD-v2, and a multi-task gain of $\textbf{1.76\%}$ on PASCAL-Context.

To assess the effectiveness and efficiency of the devised Selective Attention and Fusion Attention in the Cross-Scale Self-Attention, we implement them using a larger ViT-L encoder variant of InvPT++ and present the results in Table~\ref{tab:sa_fa_pascal} (PASCAL-Context) and Table~\ref{tab:sa_fa_nyud} (NYUD-v2). By default, we keep the top 50\% of key tokens in Selective Attention. Selective Attention achieves better performance on all tasks while utilizing much fewer FLOPs.
As a result, we choose Selective Attention as the default option for InvPT++.

Additionally, we investigate the impact of varying the proportion of selected key tokens in Selective Attention on performance. We perform experiments retaining 25\%, 50\%, and 75\% of key tokens and present the results in Table~\ref{tab:token_ratios}. We observe that maintaining a higher percentage of tokens tends to yield better performance, although it increases the computational cost.

\begin{table}[!t]
\centering
\caption{Ablation study of the InvPT++ encoders on PASCAL-Context.}
\large
\resizebox{1.\linewidth}{!}{
 \setlength{\tabcolsep}{3mm}{
    \begin{tabular}[t]{cccccc}
    \toprule
   \multirow{2}*{ \textbf{Encoder} }  & \textbf{Semseg}  & \textbf{Parsing}
       & \textbf{Saliency} & \textbf{Normal} & \textbf{Boundary}
     \\
        & mIoU $\mathbf{\uparrow}$  & mIoU $\mathbf{\uparrow}$
       & maxF $\mathbf{\uparrow}$ & mErr $\mathbf{\downarrow}$ & odsF $\mathbf{\uparrow}$
     \\
    \midrule
     Swin-T & 73.93 & 63.09 & 84.25 & 14.03 & 72.60 \\
     Swin-B  & 78.43 & 67.66 & 84.58 & 14.04 & 74.00  \\
     Swin-L & 79.65 & \textbf{69.14} & 84.78 & 14.09 & \textbf{74.80} \\
     \midrule
     Vit-B & 76.95 & 66.89 & \textbf{85.12} & \textbf{13.54} & 73.30 \\
     Vit-L & \textbf{80.22} & {69.12} & 84.74 & {13.73} & {74.20}\\
    \bottomrule
   \end{tabular}}}
\label{tab:abl_encoder_pascal}
\end{table}

\begin{table}[t]
\centering
\caption{Performance comparison of using different transformer encoder structures in InvPT++ on NYUD-v2.}
\label{tab:abl_encoder_nyud}
\resizebox{0.9\linewidth}{!}{
    \begin{tabular}{ccccccc}
    \toprule
    \multirow{2}*{ \textbf{Encoder} }   & \textbf{Semseg}  & \textbf{Depth}  & \textbf{Normal} & \textbf{Boundary} \\
         &mIoU $\mathbf{\uparrow}$ & RMSE $\mathbf{\downarrow}$ & mErr $\mathbf{\downarrow}$ & odsF $\mathbf{\uparrow}$ \\
    \midrule
     Swin-T & 44.93 & 0.5552 & 20.46 & 76.10   \\
     Swin-B & 50.98 & \textbf{0.5009} & 19.26 & 77.10 \\
     Swin-L & 52.35& 0.4921 & 18.99 & 77.90 \\
     \midrule
     Vit-B & 49.79 & 0.5318 & 18.90 & 77.10 \\
     Vit-L & \textbf{53.85} & 0.5096 & \textbf{18.67} & \textbf{78.10} \\
    \bottomrule
    \end{tabular}}
\end{table}

\begin{table*}[t]
\centering\caption{Comparison with the state-of-the-art methods on NYUD-v2 (left) and PASCAL-Context (right) datasets.
Our InvPT and InvPT++ obtain significantly superior performance than previous methods.}
% \vspace{3pt}
\label{tab:sota_s}
\begin{minipage}[c]{\textwidth}
\begin{subtable}[t]{0.48\textwidth}
\centering
\resizebox{1.\linewidth}{!}{
\setlength{\tabcolsep}{1.5mm}{
    \begin{tabular}{lcccccc}
    \toprule
         \multirow{2}*{ \textbf{Model} }   & \textbf{Semseg}  & \textbf{Depth}  & \textbf{Normal} & \textbf{Boundary} \\
         &mIoU $\mathbf{\uparrow}$ & RMSE $\mathbf{\downarrow}$ & mErr $\mathbf{\downarrow}$ & odsF $\mathbf{\uparrow}$ \\
    \midrule
    Cross-Stitch~\cite{crossstitch} & 36.34 & 0.6290 & 20.88 &76.38    \\
    PAP~\cite{papnet} & 36.72 & 0.6178 &20.82 & 76.42 \\
    PSD~\cite{psd} & 36.69 & 0.6246 & 20.87 & 76.42 \\
    PAD-Net~\cite{padnet} & 36.61 & 0.6270 & 20.85 & 76.38\\
    MTI-Net~\cite{mti} & 45.97  & 0.5365 & 20.27 & 77.86 \\
    ATRC~\cite{atrc}  & 46.33  & 0.5363 & 20.18 & 77.94 \\
    MQTransformer~\cite{xu2022multi} & 49.18 & 0.5785 & 20.81 & 77.00 \\
    \midrule
    InvPT~\cite{invpt2022} &53.56 & 0.5183 & 19.04 & 78.10\\
    InvPT++ & \textbf{53.85} & \textbf{0.5096} & \textbf{18.67} & \textbf{78.10}\\
    \bottomrule
    \end{tabular}}}
    \label{tab:sota_nyud}
\end{subtable}
\begin{subtable}[t]{0.52\textwidth}
\centering
\resizebox{1.0\linewidth}{!}{
\setlength{\tabcolsep}{1mm}{
    \begin{tabular}{lcccccc}
    \toprule
   \multirow{2}*{ \textbf{Model} }  & \textbf{Semseg}  & \textbf{Parsing}
       & \textbf{Saliency} & \textbf{Normal} & \textbf{Boundary}
     \\
        & mIoU $\mathbf{\uparrow}$  & mIoU $\mathbf{\uparrow}$
       & maxF $\mathbf{\uparrow}$ & mErr $\mathbf{\downarrow}$ & odsF $\mathbf{\uparrow}$
     \\
    \midrule
    ASTMT~\cite{astmt} & 68.00 & 61.10 & 65.70 &  14.70 & 72.40\\
    PAD-Net~\cite{padnet} & 53.60 & 59.60 & 65.80 & 15.30 & 72.50 \\
    MTI-Net~\cite{mti}  &  61.70 & 60.18 & 84.78 & 14.23 & 70.80\\
    ATRC~\cite{atrc}  & 62.69 & 59.42 & 84.70 & 14.20 & 70.96\\
    ATRC-ASPP~\cite{atrc}  & 63.60 & 60.23 & 83.91 & 14.30 & 70.86\\
    ATRC-BMTAS~\cite{atrc}  & 67.67 & 62.93 & 82.29 & 14.24 & 72.42\\
    MQTransformer~\cite{xu2022multi} & 71.25 & 60.11 & 84.05 & 14.74 & 71.80\\
    \midrule
    InvPT~\cite{invpt2022} & {79.03} & {67.61} & {84.81} & {14.15} & {73.00}\\
    InvPT++ & \textbf{80.22} & \textbf{69.12} & \textbf{84.74} & \textbf{13.73} & \textbf{74.20}\\
    \bottomrule
    \end{tabular}}}
    \label{tab:sota_pascal}
\end{subtable}
\end{minipage}
% \vspace{-20pt}
\end{table*}

\subsubsection{Ablation Study of InvPT++ Encoders}
We examine the influence of InvPT++ encoders in this section. We adopt two categories of transformer encoders, namely Swin Transformer (Swin-T, Swin-B, and Swin-L)\cite{swin} and ViT (ViT-B and ViT-L)\cite{vit}, as our InvPT++ transformer encoder. The results for PASCAL-Context are presented in Table~\ref{tab:abl_encoder_pascal}, while the outcomes for NYUD-v2 are presented in Table~\ref{tab:abl_encoder_nyud}. We notice that, within the same model family, models with higher capacity generally achieve consistent performance improvements on tasks such as Semseg and Parsing. However, the enhancement is not as apparent for other lower-level tasks (\eg, Boundary). One possible explanation for this discrepancy in performance gains could be the task competition issue during training, as discussed in previous studies~\cite{sener2018multi,kendall2018multi}.

\subsubsection{Ablation Study of Different Numbers of Stages} 
The UP-Transformer block in our approach generally comprises three stages. In Fig.~\ref{fig:stages}, we demonstrate the impact of varying the number of stages in the InvPT++ decoder on the performance of different tasks in the PASCAL-Context dataset. The model utilizes a ViT-L encoder. We observe that employing additional stages enhances the InvPT++ decoder's ability to generate improved predictions across all tasks. This is due to our InvPT++ decoder can model cross-task interaction at higher resolutions as the number of stages increases.

\subsubsection{Visualization of learned Features by InvPT++}
We provide a visualization of the learned final features from the transformer baseline (\ie, ``InvPT++ Baseline (MT)'') and our  InvPT++ model in Fig.\ref{fig:tsne_featimprove}. It further illustrates how our proposed InvPT++ model enhances the quality of the features. The statistics of the learned feature points are visualized using t-SNE\cite{tsne} on all 20 semantic classes from 1000 randomly selected samples per class in the test split of the PASCAL-Context dataset and Cityscapes dataset. It is evident that our model aids in learning more discriminative features, consequently leading to higher quantitative results. Additionally, the generated task-specific feature maps for semantic segmentation are also intuitively improved.

\subsubsection{Visualization of the Preliminary and Final Predictions}
Fig.~\ref{fig:vis_inter} presents a qualitative comparison of the preliminary and final predictions generated by InvPT++ on PASCAL-Context. We can see that the InvPT++ decoder effectively refines the preliminary predictions, resulting in significantly improved outcomes across all dense prediction tasks. 
This further verifies the effectiveness of the carefully-designed InvPT++ decoder

\begin{figure*}[t]
\centering
\includegraphics[width=1\textwidth]{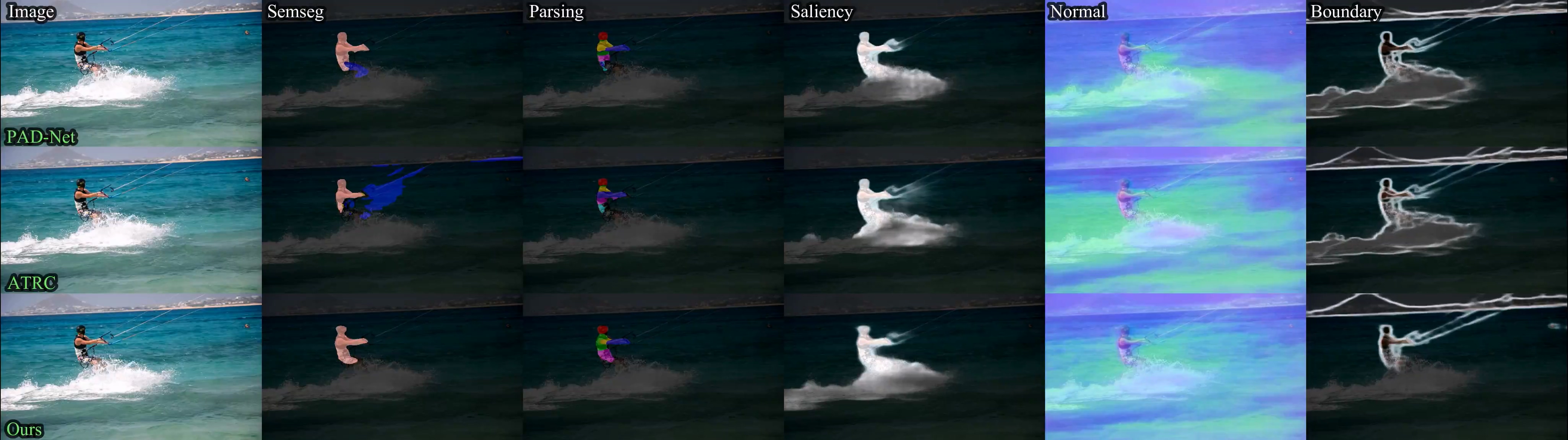}
\caption{
Visualization of the generated multi-task predictions by PAD-Net~\cite{padnet}, ATRC~\cite{atrc}, and our method on DAVIS dataset. All the models are trained with the training split of PASCAL-Context dataset. Our method yields better generalization performance on all tasks. 
}
\label{fig:demo}
\end{figure*}

\begin{table}[t]
\centering\caption{State-of-the-art comparison with previous multi-task learning methods on Cityscapes. InvPT++ significantly outperforms previous methods.}
\centering
\resizebox{.7\linewidth}{!}{
\setlength{\tabcolsep}{1.5mm}{
    \begin{tabular}{lcccccc}
    \toprule
         \multirow{2}*{ \textbf{Model} } & \multirow{2}*{ \textbf{Backbone}}  & \textbf{Semseg}  & \textbf{Depth}   \\
         & &mIoU $\mathbf{\uparrow}$ & abs Err $\mathbf{\downarrow}$  \\
    \midrule
    XTC~\cite{li2022learning} & SegNet & 74.23 & 0.0235 \\
    MTAN~\cite{liu2019MTAN} & SegNet & 75.31 & 0.0119 \\
    UR~\cite{li2022universal} & SegNet & 76.42 & 0.0117 \\
    \midrule
    PAD-Net~\cite{padnet} & ViT-B & 77.93 & 0.0102 \\
    MTI-Net~\cite{mti} & ViT-B & 78.70 & 0.0097 \\
    InvPT++ & ViT-B & \textbf{81.97} & \textbf{0.0091}\\
    \bottomrule
    \end{tabular}}}
    \label{tab:sota_cityscapes}
\end{table}

\begin{table}[t]
\centering
% \vspace{-20pt}
\caption{Joint 2D-3D Multi-task scene understanding on Cityscapes-3D. InvPT++ achieves much higher performance than the multi-task baseline model.}
\label{tab:sota_cityscapes3d}
\resizebox{0.65\linewidth}{!}{
    \begin{tabular}{clccccccccc}
    \toprule
     \multicolumn{1}{c}{ \multirow{2}*{ \textbf{Model}} }& \textbf{3Ddet} & \textbf{Semseg}  & \textbf{Disparity}\\
    \multicolumn{1}{c}{}    & mDS $\mathbf{\uparrow}$ & mIoU $\mathbf{\uparrow}$ & RMSE $\mathbf{\downarrow}$  \\
    \midrule
    Baseline & 27.49 & 67.94 & 5.33  \\
    InvPT++ & \textbf{28.57} & \textbf{74.50} & \textbf{4.86} \\
    \bottomrule
    \end{tabular}}
\end{table}

\subsection{State-of-the-art Comparison}
To showcase the outstanding performance of the proposed InvPT++ model on multi-task scene understanding, we compare it with many state-of-the-art methods. Table~\ref{tab:sota_s} provides a comparison between the proposed InvPT++ method and existing state-of-the-art techniques, including PAD-Net~\cite{padnet}, MTI-Net~\cite{mti}, ATRC~\cite{atrc}, and the transformer-based MQTransformer~\cite{xu2022multi} on both NYUD-v2 and PASCAL-Context datasets. InvPT++ consistently surpasses other methods across all nine task metrics from these two benchmarks, particularly excelling in higher-level scene understanding tasks such as Semseg and Parsing. Remarkably, on NYUD-v2, our InvPT++ surpasses  MQTransformer by~\textbf{+4.67} (mIoU) on Semseg, while on PASCAL-Context, InvPT++ outperforms MQTransformer by~\textbf{+8.97} (mIoU) and \textbf{+9.01} (mIoU) on Semseg and Parsing, respectively. 
We implement ATRC, MTI-Net, and PAD-Net with the ViT-L backbone and compare them against our InvPT++ in Table~\ref{tab:sota_pascal_vitl}. We can still observe that InvPT++ consistently outperforms the others across all tasks while consuming a reduced computational cost, providing compelling evidence of its superior effectiveness. 
We further illustrate the multi-task predictions generated by our model and the competitive ATRC~\cite{atrc} in
Fig.\ref{fig:qualitative_pascal} (for PASCAL-Context) and Fig.\ref{fig:qualitative_sota_nyud} (for NYUD-v2). The visuals clearly indicate that InvPT++ produces significantly more refined outputs. Both qualitative and quantitative results prove the superiority of our methods.

To further evaluate the effectiveness of InvPT++, we investigate its performance on another popular benchmark: Cityscapes. We implement PAD-Net and MTI-Net on Cityscapes and compare InvPT++ with several methods, including UR~\cite{li2022universal}, MTAN~\cite{liu2019MTAN}, and XTC~\cite{li2022learning}. The results are shown in Table~\ref{tab:sota_cityscapes}. InvPT++ achieves the best performance, further demonstrating the effectiveness of our proposal.

We also implement InvPT++ on the Cityscapes-3D benchmark, where it is required to simultaneously learn semantic segmentation, 3D object detection, and depth estimation. For comparison, we construct a strong baseline model that uses Swin-T as the encoder, two Conv-BN-ReLU units as task-specific decoders, and FCOS-3D~\cite{wang2021fcos3d} as the 3D detection head. The results are presented in Table~\ref{tab:sota_cityscapes3d}. InvPT++ surpasses the baseline in all tasks. Additionally, we showcase generated results on the Cityscapes-3D dataset in Fig.~\ref{fig:vis_cityscapes}, which demonstrate that InvPT++ can simultaneously generate high-quality predictions for both 2D and 3D tasks.

\subsubsection{Generalization Performance}
We further investigate our model's performance with respect to its ability to adapt to different data distributions.
We train our model using the training split of PASCAL-Context dataset and subsequently generate multi-task predictions on the DAVIS dataset. Our method is compared with PAD-Net and ATRC, and the visualizations of the outputs are displayed in Fig.~\ref{fig:demo}. Unlike other methods, our model successfully generates satisfactory predictions even when faced with an unseen data distribution. This outcome suggests that our model exhibits a significantly superior generalization capability compared to previous models.

\begin{table}[t]
% \vspace{-5pt}
\huge
\centering\caption{Comparison with the state-of-the-art methods implemented with ViT-L backbone on PASCAL-Context. Our InvPT++ demonstrates  superior performance across all tasks, utilizing the same backbone but with a reduced computational cost. }%`$\mathbf{\downarrow}$' means lower better and `$\mathbf{\uparrow}$' means higher better.}
\centering
\resizebox{1.0\linewidth}{!}{
\setlength{\tabcolsep}{1mm}{
    \begin{tabular}{l|ccccccc}
    \toprule
   \multirow{2}*{ \textbf{Model} } & \multirow{2}*{\textbf{FLOPs}} & \multirow{2}*{\textbf{\#Param}}  & \textbf{Semseg}  & \textbf{Parsing}
       & \textbf{Saliency} & \textbf{Normal} & \textbf{Boundary}
     \\
        &&& mIoU $\mathbf{\uparrow}$  & mIoU $\mathbf{\uparrow}$
       & maxF $\mathbf{\uparrow}$ & mErr $\mathbf{\downarrow}$ & odsF $\mathbf{\uparrow}$
     \\
     \midrule
    PAD-Net~\cite{padnet} &  773G   &  330M  &  78.01  &  67.12  &  79.21  &  14.37  &  72.60 \\
    MTI-Net~\cite{mti} & 774G  &   851M   &  78.31   &  67.40   &  84.75   & 14.67   & 73.00  \\
    ATRC~\cite{atrc} &871G  &  340M   &  77.11   &  66.84   &  81.20   & 14.23  &  72.10  \\
    InvPT++ & 667G & 421M &  \textbf{80.22} & \textbf{69.12} & \textbf{84.74} & \textbf{13.73} & \textbf{74.20}\\
    \bottomrule
    \end{tabular}}}
    \label{tab:sota_pascal_vitl}
\end{table}

\section{Conclusion}
\label{sec:conclusion}
This study introduces the Inverted Pyramid Multi-task Transformer (InvPT++), a potent model created for multi-task visual scene understanding. With its unique ability to model cross-task interactions within a spatially global context, InvPT++ can refine multi-task representations across diverse feature scales.
We have also proposed a new Cross-Scale Self-Attention mechanism, designed to connect attention information across different scales, thereby enhancing the modeling of cross-task interactions.
To support the learning of multi-scale features, we employ a specially designed Encoder Feature Aggregation strategy, further enhancing the effectiveness of our InvPT++ approach.
Through quantitative and qualitative experiments, we demonstrate the efficacy of various modules within our framework. The results clearly show that our method outperforms its predecessors and achieves strong state-of-the-art performances on several competitive multi-task benchmarks.

% \par \noindent \textbf{Acknowledgements.}
\section*{Acknowledgements}
This research is supported in part by the Early Career Scheme of the Research Grants Council (RGC) of the Hong Kong SAR under grant No. 26202321 and HKUST Startup Fund No. R9253.

\bibliographystyle{IEEEtran}
\bibliography{refers}


% \end{thebibliography}


\begin{IEEEbiography}
[{\includegraphics[width=1in,height=1.25in,clip,keepaspectratio]{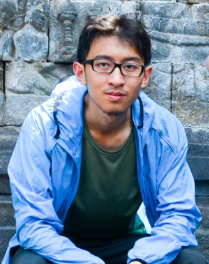}}]
{Hanrong Ye}
is currently a PhD student in the Department of Computer Science and Engineering at Hong Kong University of Science and Technology. His research interest mainly focuses on multi-modal multi-task learning.
\end{IEEEbiography}

\begin{IEEEbiography}
[{\includegraphics[width=1in,height=1.25in,clip,keepaspectratio]{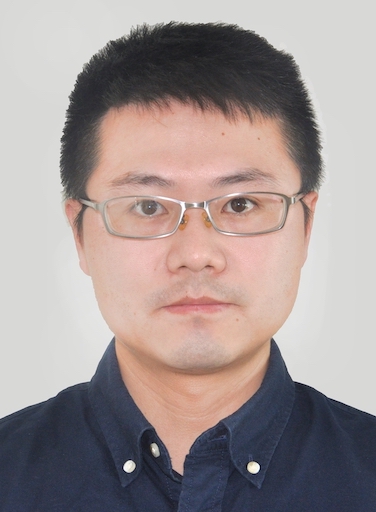}}]
{Dan Xu} is an Assistant Professor in the Department of Computer Science and Engineering at HKUST. He was a Postdoctoral Research Fellow
in VGG at the University of Oxford. He was a
Ph.D. in the Department of Computer Science at
the University of Trento. He was also a research
assistant of MM Lab at the Chinese University of
Hong Kong. He received the best scientific paper
award at ICPR 2016, and a Best Paper Nominee
at ACM MM 2018. He served as Area Chairs at multiple main-stream conferences including CVPR, AAAI, ACM Multimedia, WACV, ACCV and ICPR.
\end{IEEEbiography}

%\newpage



%\vfill


\end{document}